\RequirePackage{fix-cm}
\documentclass[twocolumn]{svjour3}          
\smartqed  
\usepackage[english]{babel}
\usepackage{graphicx}
\usepackage{amsmath}
\usepackage{amssymb}
\usepackage{mathtools}
\usepackage{float}
\usepackage[linesnumbered, ruled, vlined]{algorithm2e}
\usepackage{url}
\usepackage{bigstrut}
\usepackage{caption}
\usepackage{subcaption}
\usepackage[all]{myglossary}

\usepackage{tikz,bm,color}
\usetikzlibrary{fit}
\newcommand\addvmargin[1]{
  \node[fit=(current bounding box),inner ysep=#1,inner xsep=0]{};
}

\usepackage{array}
\usepackage{tabularx}
\usepackage{multirow}
\usepackage{longtable}
\usepackage{booktabs}

\newcolumntype{L}[1]{>{\raggedright\let\newline\\\arraybackslash\hspace{0pt}}m{#1}}
\newcolumntype{C}[1]{>{\centering\let\newline\\\arraybackslash\hspace{0pt}}m{#1}}
\newcolumntype{R}[1]{>{\raggedleft\let\newline\\\arraybackslash\hspace{0pt}}m{#1}}

\journalname{Journal of Ambient Intelligence and Humanized Computing}

\begin{document}


\title{Constrained Multi-objective Optimization for Multi-UAV Planning}

\titlerunning{Constrained Multi-objective Optimization for Multi-UAV Planning}         

\author{Cristian Ramirez-Atencia \and David Camacho \and}


\institute{Cristian Ramirez-Atencia \at
             Universidad Auton\'onoma de Madrid\\
			 \email{cristian.ramirez@inv.uam.es}
             \and
             David Camacho \at
             Universidad Auton\'onoma de Madrid\\
             \email{david.camacho@uam.es}
}

\date{Received: date / Accepted: date}

\maketitle

\begin{abstract}
Over the last decade, developments in \glspl{uav2} has greatly increased, and they are being used in many fields including surveillance, crisis management or automated mission planning. This last field implies the search of plans for missions with multiple tasks, \glspl{uav2} and \glspl{gcs}; and the optimization of several objectives, including makespan, fuel consumption or cost, among others. In this work, this problem has been solved using a \gls{moea} combined with a \gls{csp} model, which is used in the fitness function of the algorithm. The algorithm has been tested on several missions of increasing complexity, and the computational complexity of the different element considered in the missions has been studied.
\end{abstract}

\glsresetall

\section{Introduction}\label{introduction}

The potential applications of \gls{uas} are varied, including infrastructure inspection~\cite{Mathe2015Vision}, surveillance of coast borders~\cite{Turner2016UAVs} and road traffic~\cite{Chen2007Real}, disaster and crisis management~\cite{Wu2006High}, and agriculture or forestry~\cite{Merino2006Cooperative}, among others. Therefore, over the past 20 years, a large number of \gls{uas} research works have been carried out applying varied techniques and algorithms~\cite{Kendoul2012Survey}.

The systems developed to increase the autonomous capabilities of a \gls{uas} can be mostly organized into three main categories: \gls{gnc}. The \gls{gsys} is in charge of the planning and decision-making functions to achieve assigned missions or goals. Mission planning is the process of generating tactical goals, a commanding structure, coordination, and timing for a team of \glspl{uav2}. The mission plans can be generated either in advance or in real time, manually by operators or by onboard software systems, and in either centralized or distributed ways.

Currently, \glspl{uav2} are operated remotely by human operators from \glspl{gcs}, using rudimentary guidance systems, such as following preplanned or manually provided waypoints. There is an active research topic that tries to outperform guidance systems to achieve more complex tasks and missions, in order to provide a new degree of high-level autonomy to guidance systems including mission planning and decision-making capabilities.

In this work, the \gls{mpp} is modelled as a \gls{csp} and solved using a \gls{moea} optimizing several variables of the problem, such as the makespan, the cost of the mission or the risk. 

In the following section a state of the art on mission planning, \glspl{csp} and \glspl{moea} is presented. Section \ref{uavmpp} presents the \gls{mpp} aimed to solve and its modelization as a \gls{csp}. Section \ref{moeaapproach} presents the \gls{moea} designed to solve the \gls{mpp}. Then, in section \ref{experiments}, a set of experiments is performed to prove the efficiency of the \gls{moea} and show complexity of the problem as the number of variables and constraints considered increase. Finally, section \ref{conclusions} brings some conclusions and propose future works with this approach.

\section{State of the art}\label{stateoftheart}

In this section, an overview of the state of the art of the problem and techniques used in this paper is presented. First, a general introduction to the mission planning problem for \glspl{uav2} and some research studies on this topic is presented. After this, an overview of \glspl{csp} is presented, showing the different methods used in the literature to solve them. Finally, a description of \glspl{mop} and the use of \glspl{moea} to solve them is carried out.

\subsection{Mission Planning}\label{missionplanning}

Automated planning and scheduling has been an area of research in artificial intelligence for over four decades, and its application in \gls{uav2} missions is a known problem. A mission consists of as a set of goals that must be achieved by performing some tasks with a group of resources (the sensors possessed by the \glspl{uav2}) over a period of time. The whole problem can be summed up in finding the correct schedule of resource-task assignments that satisfies the proposed constraints. In addition, when several \glspl{uav2} are involved, they usually require several \glspl{gcs} for controlling all the vehicles involved. This generates a new Multi-\gls{gcs} approach that makes the problem even harder to solve.

One concept that relates with mission planning and that is sometimes confused with it, is path planning. In path planning, the vehicles are assigned a set a waypoints which compose their route, i.e. it is similar to \gls{tsp} or \gls{vrp}. On the other hand, mission planning is more related with task and resource allocation, being more similar with scheduling problems. Anyway, once mission planning is performed, path planning must also be performed on each vehicle (with their assignments) to create the final route.

The \gls{mpp} is a big challenge in actual NP-hard optimization problems. Classic planners based on graph search or logic are not suitable for this problem due to the high computational cost that the algorithms need to solve these missions. For this reason, there has been some research in the last decade proposing novel methods to solve this problem.

Sakamoto \cite{Sakamoto2006UAV} proposed a linear programming formulation of the \gls{mpp} as an extension of the \gls{vrp} with time windows for the tasks, and solved it using robust optimization. Evers et al. \cite{Evers2014Robust} presented a similar approach with robust optimization, but they extended from the orienteering problem.

Karaman et al. \cite{Karaman2011Linear} extended the \gls{vrp} with lineal temporal logic, formulated it using network flows and solved it using \gls{milp}. Fabiani et al. \cite{Fabiani2007Autonomous} modelled the problem using \gls{mdp} and solve it with dynamic programming algorithms. Another approach by Kvarnstrom et al. \cite{Kvarnstrom2010Automated} combines ideas from forward-chaining planning with partial-order planning, leading to a new hybrid \gls{pofc} framework.

Other works focus on distributed approaches for solving mission planning of a swarm of \glspl{uav2}, using the \gls{bdi} model \cite{Pascarella2013Agent}, \glspl{dddas} \cite{Madey2012Applying}, etc.

Finally, there are other approaches that formulate the \gls{mpp} as a \gls{csp} \cite{Leary2011Constrained}, using constraint techniques to solve it. Moreover, some modern approaches use bio-inspired algorithms. These methods will be discussed in the following sections.

\subsection{Constraint Satisfaction Problems}\label{csps}

A \gls{csp} \cite{Bartak1999Constraint} consists of a set of variables, each one with its own set of possible values or domain, and a set of constraints restricting the values that variables can simultaneously take. This formulation is quite appropriate for the \gls{mpp}, where a set of assignments (variables) must be done while assuring some constraints given by the different components and capabilities of the \glspl{uav2} involved.

\Glspl{csp} are usually represented as graphs where the variables are the nodes and the constraints are the edges. There exists several methods to search the space of solutions of a \gls{csp}, including \gls{bt}, \gls{bj} or \gls{fc}, among others. Most methods have a propagation phase where the constraints of the problem are checked. These methods are usually combined with consistency techniques (domain consistency, arc consistency or path consistency) to modify the \gls{csp} and ensure its local consistency conditions.

When \glspl{csp} treat with time variables (time points, time intervals or durations), as with the \gls{mpp}, these type of problems are usually called \gls{tcsp} \cite{Schwalb1998Temporal}. In these problems, the temporal constraints are characterized by the underlying set of \gls{btr}. Depending on the nature of the time variables and the constraints used, \gls{btr} can be represented using different types of algebra, such as \gls{pa}, \gls{ia} or, the most known, Allen's Interval Algebra \cite{Allen1983Maintaining}, in which the \gls{btr} is composed of the relations presented in Table \ref{tab:allen}.

\begin{table}[!h]
\caption{Allen's interval algebra}
\label{tab:allen}
\centering
\begin{tabular}{|c|c|c|}
 \hline
 Relation    &   Illustration    &   Interpretation
 \\ \hline
 $T_1 < T_2$ & \begin{tikzpicture}[baseline=0]
   \draw (0.0,0.0) -- node[above] {$T_1$} (1.0,0.0) ;
   \draw (1.5,-0.5) -- node[above] {$T_2$} (2.5,-0.5) ;
   \addvmargin{1mm}
 \end{tikzpicture} & \begin{minipage}{2cm}
 \begin{center}
 $T_1$ takes place before $T_2$
 \end{center}
 \end{minipage}
 \\ \hline
 $T_1\; m\; T_2$ & \begin{tikzpicture}[baseline=0]
   \draw (0.0,0.0) -- node[above] {$T_1$} (1.25,0.0) ;
   \draw (1.25,-0.5) -- node[above] {$T_2$} (2.5,-0.5) ;
   \addvmargin{1mm}
 \end{tikzpicture} & $T_1$ meets $T_2$
 \\ \hline
 $T_1\; o\; T_2$ & \begin{tikzpicture}[baseline=0]
   \draw (0.0,0.0) -- node[above] {$T_1$} (1.5,0.0) ;
   \draw (1.0,-0.5) -- node[above] {$T_2$} (2.5,-0.5) ;
   \addvmargin{1mm}
 \end{tikzpicture} & $T_1$ overlaps $T_2$
 \\ \hline
 $T_1\; s\; T_2$ & \begin{tikzpicture}[baseline=0]
   \draw (0.0,0.0) -- node[above] {$T_1$} (1.25,0.0) ;
   \draw (0.0,-0.5) -- node[above] {$T_2$} (2.5,-0.5) ;
   \addvmargin{1mm}
 \end{tikzpicture} & $T_1$ starts $T_2$
 \\ \hline
 $T_1\; d\; T_2$ & \begin{tikzpicture}[baseline=0]
   \draw (0.5,0.0) -- node[above] {$T_1$} (2.0,0.0) ;
   \draw (0.0,-0.5) -- node[above] {$T_2$} (2.5,-0.5) ;
   \addvmargin{1mm}
 \end{tikzpicture} & $T_1$ during $T_2$
 \\ \hline
 $T_1\; f\; T_2$ & \begin{tikzpicture}[baseline=0]
   \draw (1.25,0.0) -- node[above] {$T_1$} (2.5,0.0) ;
   \draw (0.0,-0.5) -- node[above] {$T_2$} (2.5,-0.5) ;
   \addvmargin{1mm}
 \end{tikzpicture} & $T_1$ finishes $T_2$
 \\ \hline
 $T_1 = T_2$ & \begin{tikzpicture}[baseline=0]
   \draw (0.0,0.0) -- node[above] {$T_1$} (2.5,0.0) ;
   \draw (0.0,-0.5) -- node[above] {$T_2$} (2.5,-0.5) ;
   \addvmargin{1mm}
 \end{tikzpicture} & $T_1$ is equal to $T_2$
 \\ \hline
 \end{tabular}
\end{table}

Finally, many real-life applications aim to find a good solution, and not the complete space of possible solutions. For this purpose, the \gls{csp} incorporates an optimization function, becoming a \gls{csop}. This optimization function maps every solution (complete labelling of variables) to a numerical value measuring the quality of the solution. The most widely used algorithm for finding optimal solutions is called \gls{bb}, which performs a depth first search pruning the sub-trees that exceed the bound of the best value so far. In the case of Multi-Objective Optimization, an extension of this method, known as \gls{mobb} \cite{Rodriguez-Fernandez2015Multi} can be used to find the non-dominated solutions of the problem. Other methods for solving \gls{csop} include Russian doll search, Bucket elimination, Genetic algorithms and Swarm intelligence.

\subsection{Evolutionary Algorithms and Multi-Objective Optimization}\label{ea}

In the last decades, many bio-inspired algorithms have arisen to obtain fast solutions for optimization problems. These algorithms are included in some categories, where the most known are evolutionary algorithm, swarm intelligence and artificial immune systems. Evolutionary algorithms are population-based metaheuristic optimization algorithms inspired by biological evolution, where mechanism such as reproduction, mutation, recombination or selection are imitated. The most common type of evolutionary algorithms are \glspl{ga} \cite{Holland1992Adaptation}, which are inspired by natural evolution and genetics. Other common methods are \gls{es}, \gls{de} and \gls{gp}.

\Glspl{ga} have been successfully used in many optimization problems, demonstrating to be robust and capable of finding satisfactory solutions in highly multidimensional problems. These algorithms generate an initial population of individuals (or possible solutions), where each one has a chromosome representation, which is directly related with the variables of the problem. This population is evolved during a number of generations, where each generation reproduces, crosses, mutates and selects the best individuals of the population, which are evaluated using a fitness function.

Some works have dealed with the \gls{mpp} using \glspl{ga}, such as Tian et al. \cite{Tian2006Genetic}, where they solve a reconnaissance cooperative mission aiming to minimize the travel distance and time of the vehicles. In this approach, the chromosome representation of the \gls{ga} consist of subsequences, where each one represent a reconnaissance target sequence for a \gls{uav2}. Other approach by Geng et. al. \cite{Geng2013Cooperative} proposed a graph based representation for a \gls{mpp} constrained with flight prohibited zones and enemy radar sites.

\begin{figure}[h]
		\includegraphics[width=\linewidth]{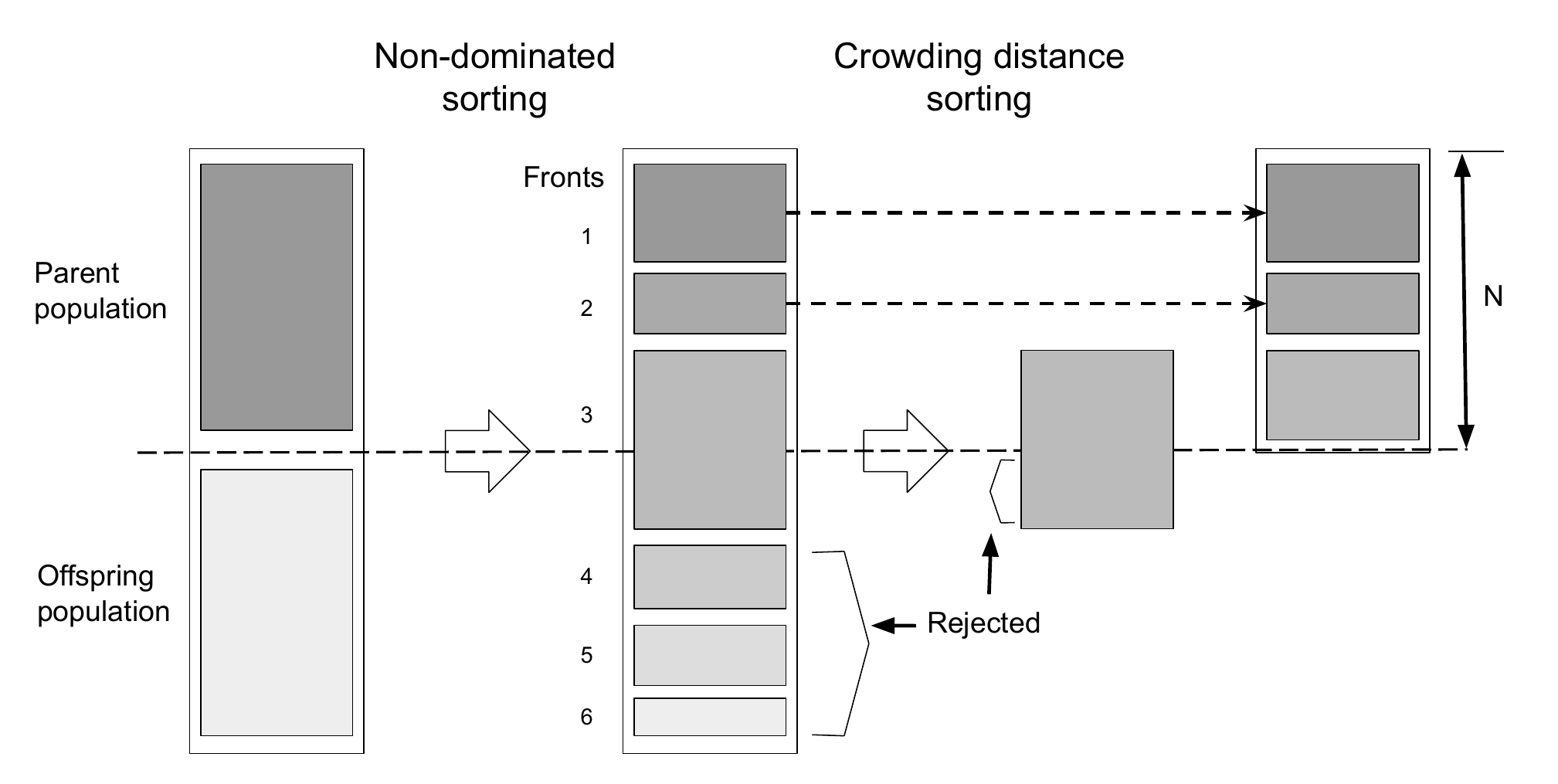}
		\centering
		\caption{NSGA-II overview of multiobjective strategies.}
		\label{fig:nsga2}
	\end{figure}

Most real-life problems, including the \gls{mpp}, aim to optimize several conflicting criteria at the same time. These \glspl{mop} have no single optimal solution that can be selected objectively; rather there is a set of solutions representing different performance trade-offs between the criteria. In these cases, algorithms usually focus on finding the \gls{pof}, i.e. the set of all non-dominated solutions of the problem. The goal of \glspl{moea} is to find non-dominated objective vectors which are as close as possible to the \gls{pof} (convergence) and evenly spread along the \gls{pof} (diversity). \gls{nsga2} has been one of the most popular algorithms over the last decade in this field \cite{Deb2002Fast}. This algorithm extends the common \gls{ga} using non-dominated ranking for the convergence and crowding distance for the diversity of the solutions (see Figure \ref{fig:nsga2}). Other popular algorithms are \gls{spea2} \cite{Zitzler2001SPEA2}, \gls{moead} \cite{Zhang2007MOEAD} and \gls{nsga3} \cite{Jain2013Improved}.

In order to evaluate the performance of \glspl{moea}, there are some metrics that can be taken into account. The hypervolume indicator \cite{Zitzler2007Hypervolume} consists of the volume enclosed by the Pareto front approximation and a reference point. The higher the hypervolume, the better the approximated \gls{pof}. Other known metrics are the \gls{gd} and the \gls{igd} \cite{Ishibuchi2015Modified}, which are based on the Euclidean distance between a solution and a reference point.

There exists some approaches of \gls{moea} for the \gls{mpp}. Pohl et al. \cite{Pohl2008Multi} compared the performance of \gls{nsga2} and \gls{spea2} for the \gls{mpp} defined as an extension of the \gls{vrp} with time windows and minimizing the path length, the number of vehicles used and the wait time. In a previous approach \cite{Ramirez-Atencia2015Hybrid}, the \gls{mpp} was considered as a task allocation problem aim to minimize the makespan and the fuel consumption of the mission. In this approach, a \gls{moga} combined with a \gls{csp} model was used, and different encodings for the order of the performance of the task were considered.

\section{Description of the Multi-UAV Mission Planning Problem}\label{uavmpp}

In this work, a more complex \gls{mpp}, with more variables and constraints, than those in the state-of-the-art is considered. As was mentioned earlier, the \gls{mpp} is modelled as a \gls{csp} \cite{Ramirez-Atencia2015Performance}. In this section, the variables and constraints are defined increasingly, starting with a simple approach and then adding more variables and constraints related with a new complexity element of the problem.

\subsection{Task allocation}
Initially, a mission is defined as a $\texttt{T}$-sized set of tasks $\mathcal{T}\doteq\{T_1,T_2,\ldots,T_\texttt{T}\}$ that must be performed by a swarm of $\texttt{U}$ \glspl{uav2} $\mathcal{U}=\{U_1,U_2,\ldots,U_\texttt{U}\}$ . Some of these tasks can be performed by several vehicles at the same time (e.g. mapping a zone or surveillance).



The first variables of the \gls{csp} that must always be considered in the \gls{mpp} are:

\begin{itemize}
\item \textbf{Task assignments} $A_\mathcal{T}$: these variables are assigned a \gls{uav2} or several if the task is Multi-UAV. These variables are represented as a binary $\texttt{T} \times \texttt{U}$ matrix. An assignment $A_\mathcal{T}[T_t,U_u]=1$ means that task $T_t$ is assigned to \gls{uav2} $U_u$.
\item \textbf{Orders} $O_\mathcal{T}$: these variables define the order in which each vehicle performs the tasks assigned to it. They are represented as an $\texttt{T} \times \texttt{U}$ matrix, where the domain for each value of the matrix is $[0 .. \texttt{T}-1]$.
\end{itemize}

Associated with these variables, there are some \textbf{order constraints} that are defined to assure that tasks performed by the same vehicle have different orders and these order values are less than the number of tasks assigned to the \gls{uav2}.

\begin{align}
\forall T_t \in \mathcal{T} \quad & \forall U_u \in \mathcal{U} \quad A_\mathcal{T}[T_t,U_u] = A_\mathcal{T}[T_{t'},U_u]=1 \nonumber \\
& \Rightarrow \quad O_\mathcal{T}[T_t,U_u]\neq O_\mathcal{T}[T_{t'},U_u] \nonumber \\
& \qquad \qquad < \sharp \left\{ { T_\tau \in \mathcal{T} }|{ A_\mathcal{T}[T_\tau,U_u]=1 } \right\}
\end{align}

\subsection{Temporal relations}
As tasks are performed within a specific time interval, there exist some \textbf{temporal constraints} that must be taken into account. These constraints require the use of some extra variables that are computed in the propagation phase of the \gls{csp}:

\begin{itemize}
\item \textbf{Departure time} $D_\mathcal{T}$ when a \gls{uav2} starts going to the task area.
\item \textbf{Path duration} $PDur_\mathcal{T}$ since the \gls{uav2} departs until it reaches the task area.
\item \textbf{Start time} $S_\mathcal{T}$ of a task.
\item \textbf{Task duration} $TDur_\mathcal{T}$. Depending on the type of task, this variable is directly provided with the mission definition (e.g. monitoring a zone) or is computed (e.g. tracking a target in a concrete path or mapping a zone using a step and stare pattern).
\item \textbf{End time} $E_\mathcal{T}$ of a task.
\item \textbf{Loiter duration} $LDur_\mathcal{T}$. This variable represents the time elapsed between the end of the previous task performed by the \gls{uav2} and the departure for the next task, which is higher than 0 when there exists some time dependencies that impede the immediate performance of the task.
\item \textbf{Return duration} $RDur_\mathcal{U}$ since the \gls{uav2} finishes its last task until it returns to the base.
\item \textbf{Return time} $R_\mathcal{U}$ of a \gls{uav2}.
\end{itemize}

With these, some temporal constraints relating each variable are considered:

\begin{align}
\forall & T_t \in \mathcal{T} \quad \forall U_u \in \mathcal{U} \quad A_\mathcal{T}[T_t,U_u] =1 \nonumber \\
& \Rightarrow \quad D_\mathcal{T}[T_t,U_u]+PDur_\mathcal{T}[T_t,U_u]=S_\mathcal{T}[T_t,U_u]
\end{align}

\begin{align}
\forall & T_t \in \mathcal{T} \quad \forall U_u \in \mathcal{U} \quad A_\mathcal{T}[T_t,U_u]=1 \nonumber \\
& \Rightarrow \quad S_\mathcal{T}[T_t,U_u]+TDur_\mathcal{T}[T_t,U_u]=E_\mathcal{T}[T_t,U_u]
\end{align}

\begin{flalign}
\forall & T_t, T_{t'} \in \mathcal{T} \quad \forall U_u \in \mathcal{U} \quad A_\mathcal{T}[T_t,U_u]=A_\mathcal{T}[T_{t'},U_u]=1 \nonumber \\
& \qquad \qquad \qquad \quad \wedge \quad O_\mathcal{T}[T_t,U_u]=O_\mathcal{T}[T_{t'},U_u]-1 \nonumber \\
& \Rightarrow \quad LDur_\mathcal{T}[T_{t'},U_u]=D_\mathcal{T}[T_{t'},U_u]-E_\mathcal{T}[T_t,U_u]
\end{flalign}

\begin{flalign}
\forall T_t \in & \mathcal{T} \quad \forall U_u \in \mathcal{U} \quad A_\mathcal{T}[T_t,U_u]=1 \nonumber \\
& \wedge \quad O_\mathcal{T}[T_t,U_u] = \sharp \left\{ { T_\tau \in \mathcal{T} }|{ A_\mathcal{T}[T_\tau,U_u]=1 } \right\}-1 \nonumber \\
& \quad \Rightarrow \quad R_\mathcal{U}[U_u]=E_\mathcal{T}[T_t,U_u]+RDur_\mathcal{T}[U_u]
\end{flalign}

In addition, it is also necessary to assure that two tasks that collide in time are never assigned to the same vehicle:

\begin{align}
\forall & T_t, T_{t'} \in \mathcal{T} \quad \forall U_u \in \mathcal{U} \quad A_\mathcal{T}[T_t,U_u]=A_\mathcal{T}[T_{t'},U_u]=1 \nonumber \\
& \qquad \qquad \qquad \qquad \qquad \wedge \quad O_\mathcal{T}[T_t,U_u] < O_\mathcal{T}[T_{t'},U_u] \nonumber \\
& \qquad \qquad \qquad \qquad \Rightarrow \quad E_\mathcal{T}[T_t,U_u] \leq D_\mathcal{T}[T_{t'},U_u]
\end{align}

On the other hand, a mission can have some task dependencies. These task dependencies trigger some \textbf{dependency constraints} in the problem. There exist two types of task dependencies: vehicle dependencies and time dependencies.

Vehicle dependencies impose if two tasks must be performed by the same or by different \glspl{uav2}, triggering the following constraints:

\begin{align}
\forall T_t, T_{t'} \in & \mathcal{T} \quad sameUAV(T_t, T_{t'}) \nonumber \\
& \Rightarrow \quad \forall U_u \in \mathcal{U} \quad A_\mathcal{T}[T_t,U_u]=A_\mathcal{T}[T_{t'},U_u]
\end{align}

\begin{align}
\forall T_t, T_{t'} \in & \mathcal{T} \quad diffUAV(T_t, T_{t'}) \nonumber \\
& \Rightarrow \quad \forall U_u \in \mathcal{U} \quad A_\mathcal{T}[T_t,U_u] \neq A_\mathcal{T}[T_{t'},U_u]
\end{align}

Time dependencies constraint the relation of the time intervals of two tasks using Allen's Interval Algebra (see Table \ref{tab:allen}). Henceforth, we assume $\forall T_t, T_{t'} \in \mathcal{T} \quad \forall U_u \in \mathcal{U} \quad A_\mathcal{T}[T_t,U_u]=A_\mathcal{T}[T_{t'},U_u]=1$, and so the following dependency constraints are defined:

\begin{equation}
T_t < T_{t'} \Rightarrow \quad E_\mathcal{T}[T_t,U_u] \le S_\mathcal{T}[T_{t'},U_u]
\end{equation}
\begin{equation}
T_t\quad m\quad T_{t'} \Rightarrow E_\mathcal{T}[T_t,U_u]=S_\mathcal{T}[T_{t'},U_u]
\end{equation}
\begin{equation}
T_t\quad o\quad T_{t'} \Rightarrow \begin{cases} S_\mathcal{T}[T_t,U_u] \le S_\mathcal{T}[T_{t'},U_u] \\ E_\mathcal{T}[T_t,U_u] \ge S_\mathcal{T}[T_{t'},U_u] \\ E_\mathcal{T}[T_t,U_u] \le E_\mathcal{T}[T_{t'},U_u] \end{cases}
\end{equation}
\begin{equation}
T_t\quad s\quad T_{t'} \Rightarrow \begin{cases} S_\mathcal{T}[T_t,U_u]=S_\mathcal{T}[T_{t'},U_u] \\ E_\mathcal{T}[T_t,U_u] \le E_\mathcal{T}[T_{t'},U_u] \end{cases}
\end{equation}
\begin{equation}
T_t\quad d\quad T_{t'} \Rightarrow \begin{cases} S_\mathcal{T}[T_t,U_u] \ge S_\mathcal{T}[T_{t'},U_u] \\ E_\mathcal{T}[T_t,U_u] \le E_\mathcal{T}[T_{t'},U_u] \end{cases}
\end{equation}
\begin{equation}
T_t\quad f\quad T_{t'} \Rightarrow \begin{cases} S_\mathcal{T}[T_t,U_u] \ge S_\mathcal{T}[T_{t'},U_u] \\ E_\mathcal{T}[T_t,U_u]=E_\mathcal{T}[T_{t'},U_u] \end{cases}
\end{equation}
\begin{equation}
T_t = T_{t'} \Rightarrow \begin{cases} S_\mathcal{T}[T_t,U_u]=S_\mathcal{T}[T_{t'},U_u] \\ E_\mathcal{T}[T_t,U_u]=E_\mathcal{T}[T_{t'},U_u] \end{cases}
\end{equation}

\subsection{Flight Profiles}
\Glspl{uav2} usually flight with a different speed and different altitude depending on the situation. For this, each vehicle have a set of flight profiles that define the speed, fuel consumption ratio and altitude of the \gls{uav2} on a path. Henceforth, the set of flight profiles of a \gls{uav2} $U_u$ is expressed as $FP(U_u)$, and given a flight profile $fp$, its speed is expressed as $s_{FP}(fp)$, its fuel consumption ratio as $f_{FP}(fp)$, and its altitude as $a_{FP}(fp)$. In this work, four flight profiles are considered, being two of them used for changes of altitude and the other two for route flight at a specific altitude:

\begin{itemize}
\item Climb profile: This profile is used by the vehicle when the altitude must be increased. Instead of the altitude, it defines the climb angle used.
\item Descent profile: This profile is used by the vehicle when the altitude must be decreased. Instead of the altitude, it defines the descent angle used.
\item Minimum consumption profile: This profile defines the speed and altitude that permit the least fuel consumption of the vehicle.
\item Maximum speed profile: This profile is used when the vehicle must flight at its maximum speed.
\end{itemize}

With these profiles, new variables of the \gls{csp} must be considered:

\begin{itemize}
	\item \textbf{Path Flight Profiles} $PFP_\mathcal{T}$: this variable sets the flight profile that the vehicle must use in its path to the task area. These variables are represented as a $\texttt{T} \times \texttt{U}$ matrix, and their domains are the flight profiles of the \gls{uav2} in the column: $PFP_\mathcal{T}[T_t,U_u \in FP(U_u)$.
	\item \textbf{Return Flight Profiles} $RFP_\mathcal{U}$: this variable sets the flight profile that the vehicle must use in its return to the base. There are $\texttt{U}$ variables of this type, and their domain is the same as the previous variables: $RFP_\mathcal{U}[U_u \in FP(U_u)$.
\end{itemize}

With these variables, and knowing the distances traversed for every path (geodesic distance), which are expressed $PDist_\mathcal{T}$ for the \textbf{Path Distance} and $RDist_\mathcal{U}$ for the \textbf{Return Distance}; it is possible to compute the path and return duration using the speed of the assigned flight profiles:

\begin{align}\label{eq:pdur}
\forall & T_t \in \mathcal{T} \quad \forall U_u \in \mathcal{U} \quad A_\mathcal{T}[T_t,U_u]=1 \nonumber \\
& \quad \Rightarrow \quad PDur_\mathcal{T}[T_t,U_u] = \frac{PDist_\mathcal{T}[T_t,U_u]}{s_{FP}(PFP_\mathcal{T}[T_t,U_u])}
\end{align}

\begin{equation}\label{eq:rdur}
\forall U_u \in \mathcal{U} \quad RDur_\mathcal{U}[U_u] = \frac{RDist_\mathcal{U}[U_u]}{s_{FP}(RFP_\mathcal{U}[U_u])}
\end{equation}

\subsection{Sensors}
The different types of tasks considered in this work require different sensors carried by the \glspl{uav2} to be performed. Henceforth, the set of sensors that a \gls{uav2} $U_u$ has are expressed as $X_\mathcal{U}$, and the set of sensors that are capable to perform a task $T_t$ are expressed as $X_\mathcal{T}(T_t)$. The sensors considered in this work are \gls{eoir} sensors, \gls{sar}, \gls{isar}, \gls{mpr} and water tanks.

In addition, the different types of task considered are defined in Table \ref{tab:tasks}.

\begin{table}[!h]
\begin{center}
\caption{Type of tasks  and the sensors capable of performing them}
\label{tab:tasks}
\resizebox{\linewidth}{!}{
\begin{tabular}{|C{2.1cm}|C{2.5cm}|C{0.8cm}|c|}
\hline
\textbf{Name} & \textbf{Description} & \textbf{Multi-UAV} & \textbf{Sensors needed} \\
\noalign{\hrule height 2pt}
Mapping a zone & Travel a zone performing a step \& stare pattern & Yes & \begin{minipage}{1in}
    \vskip 4pt
    \begin{itemize}
    	\item \gls{eoir} sensor
    \end{itemize}
    \vskip 4pt
\end{minipage}\\
\hline
Monitoring a zone & Fly circling in a zone during a specific time & No & \begin{minipage}{1in}
    \vskip 4pt
    \begin{itemize}
    	\item \gls{eoir} sensor
    \end{itemize}
    \vskip 4pt
\end{minipage}\\
\hline
Patrol a path & Follow a path & No & \begin{minipage}{1.1in}
    \vskip 4pt
    \begin{itemize}
        \item \gls{sar} radar
        \item \gls{isar} radar
    \end{itemize}
    \vskip 4pt
\end{minipage}\\
\hline
Target photographing & Go to a point and take a photo & No & \begin{minipage}{1in}
    \vskip 4pt
    \begin{itemize}
    	\item \gls{eoir} sensor
    \end{itemize}
    \vskip 4pt
\end{minipage}\\
\hline
Tracking a target & Cycle around a target during a specific time & No & \begin{minipage}{1in}
    \vskip 4pt
    \begin{itemize}
    	\item \gls{sar} radar
    	\item \gls{isar} radar
    \end{itemize}
    \vskip 4pt
\end{minipage}\\
\hline
Surveillance & Explore a zone during a specific time & Yes & \begin{minipage}{1in}
    \vskip 4pt
    \begin{itemize}
    	\item \gls{eoir} sensor
    	\item \gls{sar} radar
    	\item \gls{isar} radar
    	\item \gls{mpr} radar
    \end{itemize}
    \vskip 4pt
\end{minipage}\\
\hline
Fire Extinguishing & Put out a fire in a point & No & \begin{minipage}{1in}
    \vskip 4pt
    \begin{itemize}
    	\item Water tank
    \end{itemize}
    \vskip 4pt
\end{minipage}\\
\hline
\end{tabular}
}
\end{center}
\end{table}

Each sensor has specific values of speed and altitude for its optimal use. Henceforth, the optimal speed of a sensor $x$ is expressed as $s_X(x)$, and the optimal altitude is expressed as $a_X(x)$.

When a \gls{uav2} has more than one sensor capable of performing a task assigned to it, it is necessary to select a specific sensor for performing that task. For this, new variables of the \gls{csp} are defined: the \textbf{Task Sensors} $TX_\mathcal{T}$, which set the sensor of the \gls{uav2} used for the task performance. These variables are represented as a $\texttt{T} \times \texttt{U}$ matrix, being the different sensors their domain.

These sensor assignments come hand in hand with some \textbf{Sensor constraints}. These constraints check that the \gls{uav2} has the sensors assigned to perform the task:

\begin{align}
\forall & T_t \in \mathcal{T} \quad \forall U_u \in \mathcal{U} \quad A_\mathcal{T}[T_t,U_u]=1 \nonumber \\
& \qquad \Rightarrow \quad TX_\mathcal{T}[T_t,U_u] \in X_\mathcal{U}(U_u) \cap X_\mathcal{T}(T_t) \neq \emptyset
\end{align}

\subsection{GCS assignments}
When there are many \glspl{uav2} in a mission, most times an only \gls{gcs} is not capable of controlling all of them. In this approach, a number $\texttt{G}$ of \glspl{gcs} $\mathcal{G}\doteq\{G_1,G_2,\ldots,G_\texttt{G}\}$ are considered for controlling the \glspl{uav2}.

This consideration leads to the last variables of the \gls{csp}: the \textbf{GCS assignments} $A_\mathcal{G}$. There are $\texttt{U}$ variables of this type, one per vehicle, and their domain is $[-1 .. \texttt{G}-1]$, where $-1$ is only assigned when the \gls{uav2} does not perform any task.

Each \gls{gcs} $G_g$ supports a maximum number of vehicles being controlled at the same time $maxU_\mathcal{G}(G_g)$, and is only capable of controlling some specific types of \gls{uav2} $typeU_\mathcal{G}(G_g)$. In addition, it has a within range of communications $wr_\mathcal{G}(G_g)$, being not able of control the vehicles out of this range.

With these features, some \textbf{GCS constraints} have to be considered. These constraints assure that the \glspl{uav2} assigned to the \gls{gcs} are of a supported type, and the maximum number of vehicles that the \gls{gcs} can handle is not overpassed:

\begin{align}
\forall U_u \in \mathcal{U} \quad \forall G_g \in \mathcal{G} & \quad A_\mathcal{G}[U_u]=G_g \nonumber \\
& \Rightarrow \quad type(U_u) \subset typeU_\mathcal{G}(G_g)
\end{align}

\begin{equation}
\forall G_g \in \mathcal{G} \; \; \sharp \left\{ { U_u\in \mathcal{U} }|{ A_\mathcal{G}[U_u]=G_g } \right\} < maxU_\mathcal{G}(G_g)
\end{equation}

Finally, in order to check that the \glspl{uav2} are inside the range of the \gls{gcs} at every moment, we define the position $Pos_\mathcal{G}(G_g)$ of the \gls{gcs} $G_g$, and the position $Pos_\mathcal{U_u,\tau}$ of the \gls{uav2} $U_u$ at time $\tau$. Also, the geodesic distance function $GeoDist(p_1,p_2)$ is defined for each pair of (Latitude, Longitude, Altitude) points $p_1, p_2$. With this, the \gls{gcs} constraint is defined:

\begin{align}
\forall & U_u \in \mathcal{U} \quad \forall G_g \in \mathcal{G} \quad A_\mathcal{G}[U_u]=G_g \quad \Rightarrow \quad \forall \tau \in \mathbb{R} \nonumber \\
& \quad GeoDist(Pos_\mathcal{U}(U_u,\tau),Pos_\mathcal{G}(G_g)) \leq wr_\mathcal{G}(G_g)
\end{align}

\subsection{Checking constraints}
The \glspl{uav2} of a mission have some features that must be considered when checking if a plan is correct. These features include the already explained sensors and flight profiles of the vehicle, and also the type of the vehicle. In this work, the types considered are \gls{urav}, \gls{male}, \gls{hale} and \gls{ucav}, which are described in Table \ref{tab:uavs}.

Apart from these, there are some features that constraint the use of the vehicle: the maximum flight time $maxFT_\mathcal{U}(U_u)$ that the \gls{uav2} can stay in flight, the maximum distance $maxDist_\mathcal{U}(U_u)$ that the \gls{uav2} can traverse during a mission, the cost per hour $cost_\mathcal{U}(U_u)$ of the \gls{uav2}, the maximum speed $maxS_\mathcal{U}(U_u)$ that the \gls{uav2} can attain, the maximum altitude $maxA_\mathcal{U}(U_u)$ that the \gls{uav2} can reach in flight and the maximum fuel $maxF_\mathcal{U}(U_u)$ in the \gls{uav2} tank.

Other features that must be considered are the initial settings of the \gls{uav2} $U_u$ at the start of the mission: its initial position $Pos_\mathcal{U}(U_u)$ and its initial amount of fuel $F_\mathcal{U}(U_u)$.

\begin{table*}
\begin{center}
\caption{Different types of UAVs considered and their features.}
\label{tab:uavs}
\begin{tabular}{|c|C{1.5cm}|C{1.7cm}|c|C{1.7cm}|C{2cm}|C{1.5cm}|c|}
\hline
\textbf{Name} & \textbf{Max. Distance (NM)} & \textbf{Max. Flight Time (h)} & \textbf{Cost/h} & \textbf{Max. Speed (kt)} & \textbf{Max. Altitude (ft)} & \textbf{Max. Fuel (kg)} & \textbf{Available Sensors} \\
\noalign{\hrule height 2pt}
\textbf{URAV} & 1000 & 20 & 5 & 120 & 20000 & 500 & \begin{minipage}{1in}
    \vskip 4pt
    \begin{itemize}
    	\item \gls{eoir} sensor
    	\item Water tank
    \end{itemize}
    \vskip 4pt
\end{minipage}\\
\hline
\textbf{MALE}  & 5000 & 30 & 10 & 250 & 40000 & 2500 & \begin{minipage}{1.1in}
    \vskip 4pt
    \begin{itemize}
    	\item \gls{eoir} sensor
        \item \gls{mpr} radar
    \end{itemize}
    \vskip 4pt
\end{minipage}\\
\hline
\textbf{HALE}  & 15000 & 40 & 15 & 400 & 65000 & 6000 & \begin{minipage}{1in}
    \vskip 4pt
    \begin{itemize}
    	\item \gls{eoir} sensor
        \item \gls{isar} radar
    \end{itemize}
    \vskip 4pt
\end{minipage}\\
\hline
\textbf{UCAV}  & 1500 & 15 & 25 & 450 & 35000 & 9000 & \begin{minipage}{1in}
    \vskip 4pt
    \begin{itemize}
        \item \gls{sar} radar
    \end{itemize}
    \vskip 4pt
\end{minipage}\\
\hline
\end{tabular}
\end{center}
\end{table*}

With these, there are some \textbf{Checking constraints} that must be met to assure the correct functioning of the \gls{uav2}.

\begin{itemize}
\item Flight time constraints: they assure that the total flight time for each vehicle is less than its maximum:

\begin{align}
& \forall U_u \in \mathcal{U} \quad FT[U_u]= \sum _{\mathclap{\substack{T_t\in \mathcal{T}\\ A_\mathcal{T}[T_t,U_u]=1}}}{ (PDur_\mathcal{T}[T_t,U_u] } \nonumber \\
& +TDur_\mathcal{T}[T_t,U_u]+LDur_\mathcal{T}[T_t,U_u]) + RDur_\mathcal{U}[U_u] \nonumber \\
& \qquad \qquad \qquad \qquad \qquad \qquad < maxFT_\mathcal{U}(U_u)
\end{align}

\item Distance constraints: they assure that the distance traversed by each vehicle is less than its maximum. To compute these constraints, the 
\textbf{Path Distance} $PDist_\mathcal{T}$, the \textbf{Task Distance} $TDist_\mathcal{T}$, the \textbf{Loiter Distance} $LDist_\mathcal{T}$ and the \textbf{Return Distance} $RDist_\mathcal{U}$ are needed.

\begin{align}
& \forall U_u \in \mathcal{U} \quad Dist[U_u]= \sum _{\mathclap{\substack{T_t\in \mathcal{T}\\ A_\mathcal{T}[T_t,U_u]=1}}}{ (PDist_\mathcal{T}[T_t,U_u] } \nonumber \\
& +TDist_\mathcal{T}[T_t,U_u]+LDist_\mathcal{T}[T_t,U_u]) + RDist_\mathcal{U}[U_u] \nonumber \\
& \qquad \qquad \qquad \qquad \qquad \qquad < maxDist_\mathcal{U}(U_u)
\end{align}

Path and return distance are computed as the sums of distances between the points of the path employed in each case. Task distance is computed similarly in the case of tasks with no specific duration. If the task has a specific duration, then the task distance of task $T_t$ performed by \gls{uav2} $U_u$ can be computed using this duration and the optimal speed given by the sensor used:

\begin{equation}
TDist_\mathcal{T}[T_t,U_u] = TDur_\mathcal{T}[T_t,U_u] \times s_{X}(TX_\mathcal{T}[T_t,U_u])
\end{equation}

When a loiter is needed, the minimum consumption flight profile is used. Then the loiter distance of task $T_t$ performed by \gls{uav2} $U_u$ can be computed as:

\begin{equation}
LDist_\mathcal{T}[T_t,U_u] = LDur_\mathcal{T}[T_t,U_u] \times s_{FP}(MINC(U_u))
\end{equation}

\item Fuel constraints: they assure that the fuel consumed by each \gls{uav2} is less than its initial fuel. To compute these constraints, the \textbf{Path Fuel Consumption} $PF_\mathcal{T}$, the \textbf{Task Fuel Consumption} $TF_\mathcal{T}$, the \textbf{Loiter Fuel Consumption} $LF_\mathcal{T}$ and the \textbf{Return Fuel Consumption} $RF_\mathcal{U}$ are needed.

\begin{align}
& \forall U_u \in \mathcal{U} \quad F[U_u]= \sum _{\mathclap{\substack{T_t\in \mathcal{T}\\ A_\mathcal{T}[T_t,U_u]=1}}}{ (PF_\mathcal{T}[T_t,U_u] + TF_\mathcal{T}[T_t,U_u]} \nonumber \\
& \qquad \quad \; +LF_\mathcal{T}[T_t,U_u]) + RF_\mathcal{U}[U_u] < F_\mathcal{U}(U_u)
\end{align}

Each one of these fuel consumptions is computed as the product of its associated duration and the fuel consumption ratio specified by the flight profile or the sensor, or the minimum consumption in the case of loiter.

When checking the fuel consumption for each \gls{uav2}, a risk can be considered when the remaining amount of fuel at the end of the mission is low. For this, the remaining fuel is used along with a risked fuel usage value provided by the operator to compute the risk of the mission.

\end{itemize}

\subsection{Path Constraints}
\begin{figure}[!h]
	\includegraphics[width=\linewidth]{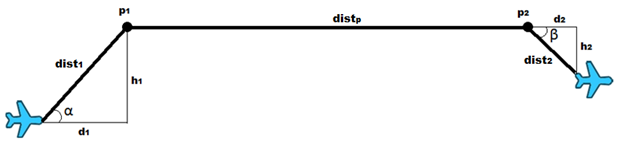}
	\centering
	\caption{Climb/Descend situation in path computing.}
	\label{fig:climbdescent}
\end{figure}

In realistic \gls{uav2} \glspl{mpp}, paths are not usually a straight line, but instead \glspl{uav2} have to perform some climb or descent in order to reach the altitude specified by the flight profile or the optimal altitude for the use of the sensor. In the case of both the path of the \gls{uav2} to the task area and the return to the base, the \gls{uav2} departs with an initial altitude that could be bigger than the flight profile altitude (and so a descent must be done) or smaller than it (so a climb must be done). So the climb or descent is performed, the \gls{uav2} flies heading to the task area (or the base) and before arriving, another change of altitude is needed to reach the optimal altitude of the sensor used in the task (or the terrain altitude in order to land). This situation is represented in Figure \ref{fig:climbdescent}.

On the other hand, some missions may have \glspl{nfz} and terrain obstacles (a \textit{DTED} file is used to define the terrain) that could be in the trajectory of the \gls{uav2} and that must be avoided. For this, Theta* \cite{Nash2007Theta} is used, returning a set of waypoints which define the route. These computations are time consuming, so when they are used, the runtime may increase significantly.

So now, the distance, duration and fuel consumption variables defined previously (see equations \ref{eq:pdur} and \ref{eq:rdur}) must be updated considering the climb and descent performed, and the waypoints obtained by Theta*.

On the other hand, when considering these more realistic paths, it is necessary to take into account the distance of the \glspl{uav2} to the ground, in order to avoid collision risks. So, for each \gls{uav2} $U_u$, the minimum distance to ground $minDG_\mathcal{U}[U_u]$ is computed. This variable will be used along with a risked distance to ground value provided by the operator to compute the risk of the mission.

Similarly, to avoid collision risks, the distance between \glspl{uav2}  must also be computed. With the time route for each pair of vehicles, the position of both vehicles at the same time is checked, storing the minimum distance obtained so far. With this, the minimum distance $minD_\mathcal{U}(U_u,U_{u'})$ between two vehicles $U_u$ and $U_{u'}$ is obtained. Similarly as with the minimum distance to ground, a risked distance between \glspl{uav2} value is provided by the operator, and the risk of the mission is updated.

\subsection{LOS constraints}
Apart from checking the within range of a \gls{gcs}, it is also necessary to assure that the \gls{gcs} and the \gls{uav2} being controlled are in \gls{los}. This means that there is no obstacle in the line of communications joining them. These \textbf{\gls{los} constraints} are quite important when missions are carried out in mountainous terrains.

\begin{align}
\forall & U_u \in \mathcal{U} \quad \forall G_g \in \mathcal{G} \quad A_\mathcal{G}[U_u]=G_g \nonumber \\
& \quad \Rightarrow \quad \forall \tau \in \mathbb{R} \quad  LOS(Pos_\mathcal{U}(U_u,\tau),Pos_\mathcal{G}(G_g))
\end{align}

The $LOS$ function used to compute the line of sight between two points, uses the \textit{DTED} terrain file to check if there any terrain in the line joining both points. This function is quite time consuming, so the runtime of the algorithm used may increase significantly when considering these constraints.

\subsection{Optimization variables}\label{objectives}
Apart from the variables and constraints considered in the \gls{csp} mode, the \gls{mpp} is a \gls{mop}, and several objectives must be considered when optimizing the problem:

\begin{enumerate}
\item The makespan or time when the mission ends (all the vehicles have returned):

\begin{equation}
M = \max _{ U_u \in \mathcal{U} }{ R_\mathcal{U}(U_u) }
\end{equation}

\item The total cost of the mission, computed as the sum of the individual cost of each \gls{uav2}:

\begin{equation}
C = \sum _{ U_u \in \mathcal{U} }{ cost_\mathcal{U}(U_u) \times FT[U_u] }
\end{equation}

\item The risk of the mission, computed as an average of risk percentages of low distance to ground $R_{Ground}$, low distance between \glspl{uav2} $R_{UAVs}$ and low fuel $R_{Fuel}$ at the end of the mission.

\begin{equation}
R = \frac{R_{Ground} + R_{UAVs} + R_{Fuel}}{3}
\end{equation}

\item The number of \glspl{uav2} used in the mission:

\begin{equation}
N_\mathcal{U} = \sharp \left\{ { U_u\in \mathcal{U} }|{ \exists T_t \in \mathcal{T} \quad A_\mathcal{T}[T_t,U_u]=1 } \right\}
\end{equation}

\item The total fuel consumed by all \glspl{uav2} during the mission:

\begin{equation}
F = \sum _{ U_u\in \mathcal{U} }{ F[U_u] }
\end{equation}

\item The total flight time of all \glspl{uav2} during the mission:

\begin{equation}
FT = \sum _{ U_u\in \mathcal{U} }{ FT[U_u] }
\end{equation}

\end{enumerate}

\section{MOEA-CSP Algorithm for Multi-UAV Mission Planning Problems}\label{moeaapproach}
The \gls{mpp} designed in the previous section is a quite complex problem with a large amount of constraints involved, and a huge number of solutions may be generated with classic methods. For this, and to the multi-objective nature of the problem, a \gls{moea} has been used to solve this problem. In this approach, an extension of \gls{nsga2} is proposed. A \gls{csp} model is used as a penalty function in the evaluation phase of the algorithm. In a second approach, in order to increase the convergence of the problem by reducing the search space, some of the constraints of the \gls{csp} are considered in the setup and operators of the algorithm.

This section describes this algorithm, including the encoding, the fitness function designed, the genetic operators implemented and the constraints considered in the assignment and mutation.

\subsection{Encoding}
The encoding of the \gls{mpp} has been based on the variables of the \gls{csp} model, taking into account the chromosome representation of the orders presented in a previous work \cite{Ramirez-Atencia2015Hybrid}. This encoding consist of six alleles (see Figure \ref{fig:chromosome}):

	\begin{figure*}[!h]
		\includegraphics[width=0.8\textwidth]{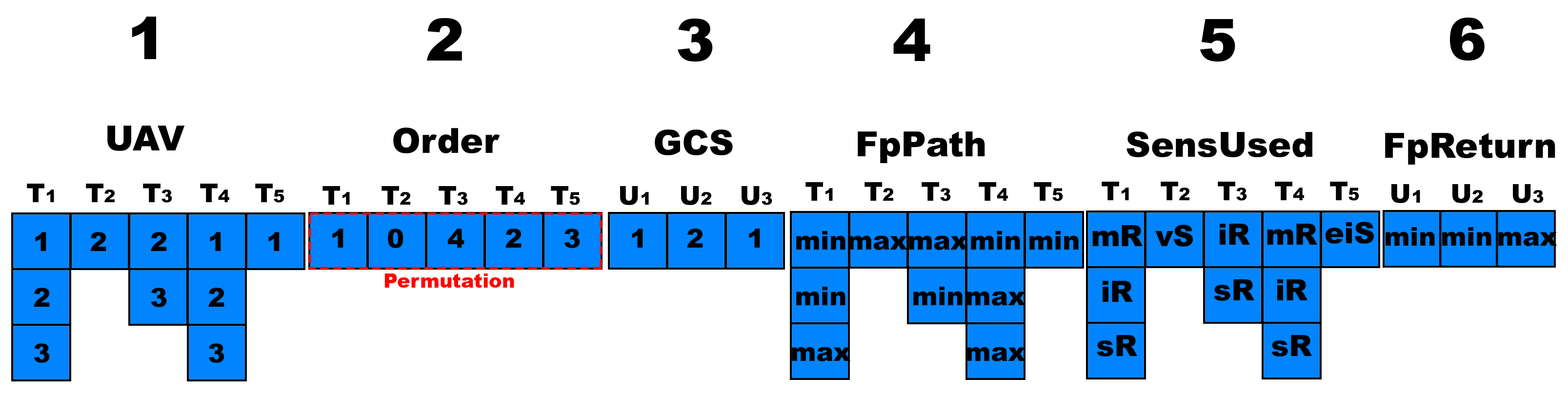}
		\centering
		\caption{Example of an individual that represents a possible solution for a problem with 5 tasks, 3 UAVs and 2 GCSs.}
		\label{fig:chromosome}
	\end{figure*}

\begin{enumerate}
	\item \gls{uav2} assignments for each task. Based on the task assignments, each cell contains the identifier of the \gls{uav2} assigned to that task. When some task is Multi-UAV, its cell contains a vector representing the different vehicles assigned to it. The number of combinations, or space size, for this allele is, considering $\texttt{T}_{MU}$ the number of Multi-UAV tasks, $\texttt{U}^{\texttt{T}-\texttt{T}_{MU}} \times (2^{\texttt{U}}-1)^{\texttt{T}_{MU}}$.
	\item Permutation of the task orders. These values indicate the absolute order of the tasks. When several tasks are assigned to the same vehicle, the ones with a lower order value start first, while the tasks with higher values go last. The space size for this allele is $\texttt{T}!$.
	\item \glspl{gcs} controlling each \gls{uav2}. This is the same as the \gls{gcs} assignments of the \gls{csp} model. The space size for this allele is $\texttt{G}^{\texttt{U}}$.
	\item Flight profiles used for each \gls{uav2} in the path to the task area. The flight profiles considered in this problem are the minimum consumption profile ($min$) and the maximum speed profile ($max$). As in the first allele, some of the cells could contain a vector if the corresponding task is performed by several vehicles. The space size for this allele is $2^{\texttt{T}-\texttt{T}_{MU}} \times (2^{\texttt{U}+1}-2)^{\texttt{T}_{MU}}$.
	\item Sensors used for the task performance by each \gls{uav2}. The values considered in this approach are eiS (\gls{eoir} sensor), sR (\gls{sar} radar), iR(\gls{isar} radar), mR (\gls{mpr} radar) and wt (water tank). Some of the cells contain a vector because of the Multi-UAV nature of the task. Considering the worst where all vehicles have all $s$ sensors, then the space size is bounded by $s^{\texttt{T}-\texttt{T}_{MU}} \times (\sum_{i=1}^{\texttt{U}} s^i)^{\texttt{T}_{MU}}$.
	\item Flight Profiles used by each \gls{uav2} to return to the base. This allele corresponds to the Return Flight Profiles variable of the \gls{csp} model. The space size for this allele is $2^{\texttt{U}}$.  
\end{enumerate}

Multiplying the different sizes for each allele and reducing the resultant formula, the total space size of the problem is as follows:

\begin{align}
\texttt{T}! \times \texttt{U}^{\texttt{T}-\texttt{T}_{MU}} \times \texttt{G}^{\texttt{U}} & \times 2^{\texttt{T}+\texttt{U}} \times (2^{\texttt{U}} - 1)^{2 \times \texttt{T}_{MU}} \times s^{\texttt{T}} \nonumber \\
& \quad \; \times (s^{\texttt{U}}-1)^{\texttt{T}_{MU}} \times (s-1)^{-\texttt{T}_{MU}}
\end{align}

\subsection{Fitness Function}\label{fitness}
The evaluation phase of the \gls{moea} is performed using a fitness function composed of two steps:

\begin{enumerate}
\item Given a specific solution, the \gls{csp} model handles that all constraints are fulfilled. If not, it returns the number of unfulfilled constraints, and these solutions are not considered when other valid solutions were found. For complex problems, valid solutions take much runtime as all constraints are checked.
\item If all constraints are fulfilled, a multi-objective function minimizing the objectives of the problem is applied. The objectives considered were presented in section \ref{objectives}.
\end{enumerate}

In the case when all the individuals of the population are invalid, those with the fewer number of invalid constraints are selected for reproduction.

As in \gls{nsga2} approach, after the evaluation of the solutions, the Pareto front is extracted and the solutions are ordered according to the crowding distance.

\subsection{Genetic Operators}\label{operators}
As the encoding presented has been specifically developed for this problem, so are the operators used in the algorithm.
The crossover operator developed is used to combine the chromosomes of each pair of parents to generate a new pair of children. This operator applies a specific crossover operation at each of the alleles. The first, fourth and fifth allele, i.e. the task assignments, path flight profiles and sensor assignments, are applied a 2-point crossover, using the same cross points in all alleles so the size for Multi-\gls{uav} tasks is maintained. The second allele, as it is a permutation, is applied a \gls{pmx}. This passes a chunk of values from one parent to the other and then performs a replacement of the invalid values of the new child based on its previous parent. Finally, the third and sixth alleles are applied another 2-point crossover with different points. 
 

Once the new pair of individuals has been generated from crossover operation, a mutation operator is required. This genetic operator helps to avoid that the obtained solutions stagnate at local minimums. This mutation operator is designed to perform a uniform mutation over the same genes for the first, fourth and fifth allele in order to maintain the size of Multi-\gls{uav} tasks. On the other hand, the second allele is applied an Insert Mutation, which will select two random positions from the permutation and move the second one next to the first one. Finally, the third and sixth allele are updated using another uniform mutation. 


\subsection{Adding constraints to the \gls{ga} process}\label{gaconstraints}
In order to facilitate the convergence of the problem, some constraints have been taken from the \gls{csp} model and added into to the \gls{ga} setup, crossover and mutation operators. This way, the algorithm will avoid assigning some invalid solutions due to these constraints, and therefore the search space is reduced. The constraints considered have been:

\begin{itemize}
\item \textbf{Sensor constraints}: As a sensor must be assigned after the assignment of a \gls{uav2} to a task, it is pretty convenient to avoid selecting the sensors required by the task that are not available by the \gls{uav2}. The setup and the mutation operation have been changed so the domain for each sensor gene is updated, after the task assignments gene are set, to $X_\mathcal{U}(U_u) \cap X_\mathcal{T}(T_t)$.
\item \textbf{\gls{gcs} constraints}: When assigning a \gls{uav2} to the \gls{gcs} controlling it, it is possible to check that this \gls{gcs} can control this type of \gls{uav2}. So the domain of the \gls{gcs} gene can be reduced to the \glspl{gcs} capable of controlling that type of vehicle. On the other hand, once the entire \gls{gcs} assignments have been done, it is possible to check, both in the setup, crossover and mutation, if every \gls{gcs} is not assigned more \glspl{uav2} than the maximum it can handle.
\item \textbf{Dependency constraints}: \gls{uav} dependencies are directly checked when task assignments are done in setup and mutation. If two task must be assigned the same vehicle, the second task is directly assigned the same \gls{uav}; and if two task must be performed by different \glspl{uav}, the domain of the second one is reduced by deleting the \gls{uav} assigned to the first one. Moreover, the Path Consistency algorithm used to check the time dependencies, so order assignments are correct, can be used in the \gls{ga} setup and operators before generating a new solution.
\end{itemize}

\subsection{Algorithm}
The new approach is presented in Algorithm \ref{alg:moga}. First, an initial population is randomly generated (Line 1) taking into account the constraints of section \ref{gaconstraints}. Then, the evaluation of the individuals is performed using the fitness function (Lines 6-14) explained in section \ref{fitness}.

\begin{algorithm}[h]
 \caption{NSGA-II with CSP model for Mission Planning Problems}
 \label{alg:moga}
\DontPrintSemicolon
\KwIn{ A mission problem $P$. The set of $m$-objectives $O$. And positive numbers elitism $\mu$, population size $\lambda$, $mutprobability$, stopping criteria limit $stopGen$ and maximum number of generations $maxGen$}
\KwOut{Pareto Front obtained with best solutions}
 $S \gets$ randomly generated set of $\lambda$ individuals\;
 $i \gets 1$\;
 $convergence \gets 0$ \;
 $pof \gets \emptyset$ \;
 \While{$i \leq maxGen \land convergence < stopGen$}{
  	\For{$j \gets 1$ \textbf{to} $|S|$}{
        $[valid,numInvalidC] \gets CSP_{Check}(S_j)$\;
        $Fit \gets new Fitness()$\;
        $Fit.valid \gets valid$\;
        \If {$valid$}{
    	    $Fit.obj \gets MultiObjectiveFitness(S_j, O)$ \;
        }
        \Else {
    	    $Fit.numInvalid \gets numInvalidC)$ \;
        }
        $S_j.Fit \gets Fit$\;
    }
	$S \gets buildNSGA2Archive(S, \lambda)$ \;    
    $newpof \gets createPOF(S)$ \;
    \If {$newpof = pof$}{
    	$convergence \gets convergence + 1$ \;
    }
    $pof \gets newpof$\;
    $newS \gets SelectElites(S, \mu)$ \;
    \For{$j \gets \mu$ \textbf{to} $\lambda$}{
      $p1,p2 \gets TournamentSelection(S$) \;
      $i1,i2 \gets Crossover(p1, p2)$ \;
      $i1 \gets Mutation(i1,mutprobability)$ \;
      $i2 \gets Mutation(i2,mutprobability)$ \;
      $newS \gets newS \cup \{i1, i2\}$ \;
    }
	$S \gets S \cup newS$ \;    
    $i \gets i + 1$ \; 
 }
 \Return{pof}\;
\end{algorithm}

After that, the \textit{buildArchive} process of \gls{nsga2} (Line 15) is performed, generating a set of  ranked vectors, where the first one represents the Pareto front of the solutions ranked by crowding distance. Then, an elitist selection is performed, where the $\mu$ best individuals in the population are retained (Line 20). Afterwards, a tournament selection over these $\mu$ individuals (Line 22) selects those that will be applied the crossover (Line 23) and mutation (Lines 24-25) operators explained in section \ref{operators} considering the constraints of section \ref{gaconstraints}.

Finally, the stopping criteria designed for this algorithm compares the Pareto front obtained so far in each generation with the Pareto front from the previous generations (Lines 17-19). The algorithm stops when the front remains unchanged during a specific number of generation $stopGen$.

\section{Experiments}\label{experiments}
In this section we explain the experiments carried out to test the functionality of the new \gls{nsga2}-\gls{csp} approach for \gls{mpp}.

The first experiment studies the performance of the algorithm considering different variables of the \gls{csp} model in the \gls{mpp} (e.g. a first approach only takes into account task assignments and orders, another approach adds flight profiles, another adds \gls{gcs} assignments, etc.).

Then, a second experiment studies the performance of the algorithm with an increasing number of constraints considered (e.g. a first approach only considers temporal constraints, then checking constraints are added, then path constraints, etc.).

Finally, the last experiments compares the performance of the algorithm when no constraints are checked inside the \gls{ga} process against the final approach considering the sensor, \gls{gcs} and dependency constraints in the setup and genetic operators.

\begin{figure*}[!t]
    \centering
    \begin{subfigure}[b]{0.31\textwidth}
        \includegraphics[width=\textwidth]{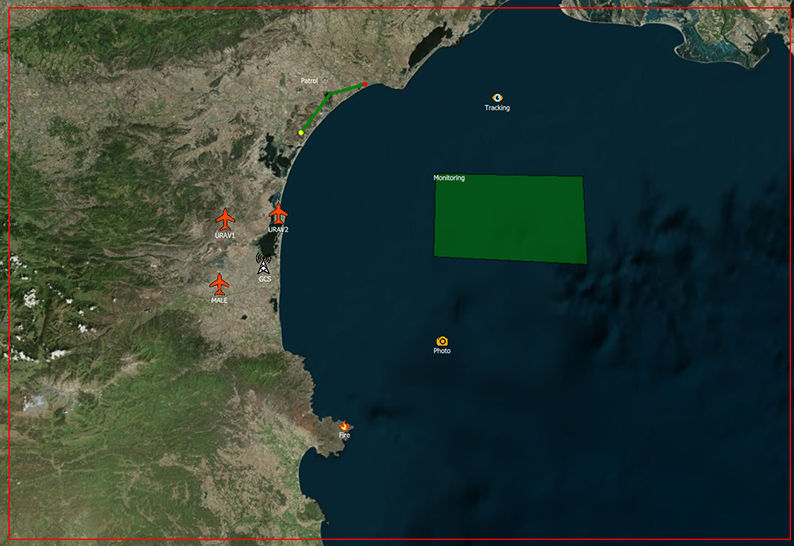}
        \caption{Mission 1.}
        \label{fig:d11}
    \end{subfigure}
    \quad
    \begin{subfigure}[b]{0.31\textwidth}
        \includegraphics[width=\textwidth]{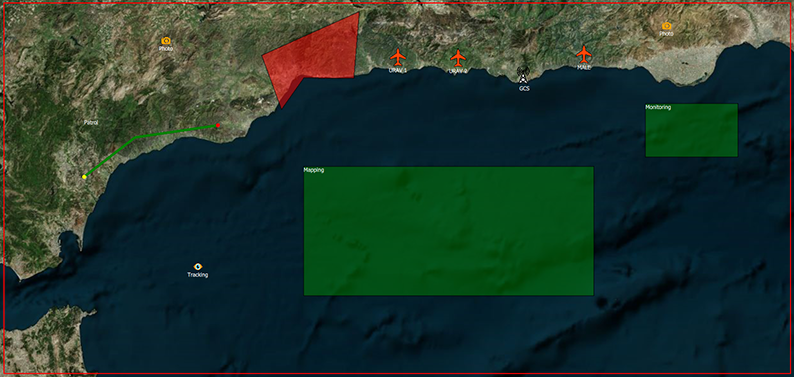}
        \caption{Mission 2.}
        \label{fig:d12}
    \end{subfigure}
    \quad
    \begin{subfigure}[b]{0.31\textwidth}
        \includegraphics[width=\textwidth]{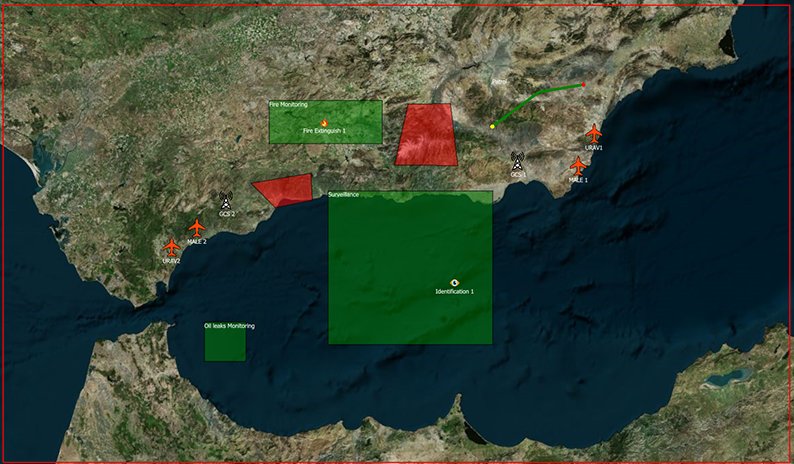}
        \caption{Mission 3.}
        \label{fig:d13}
    \end{subfigure}
    \quad
    \begin{subfigure}[b]{0.31\textwidth}
        \includegraphics[width=\textwidth]{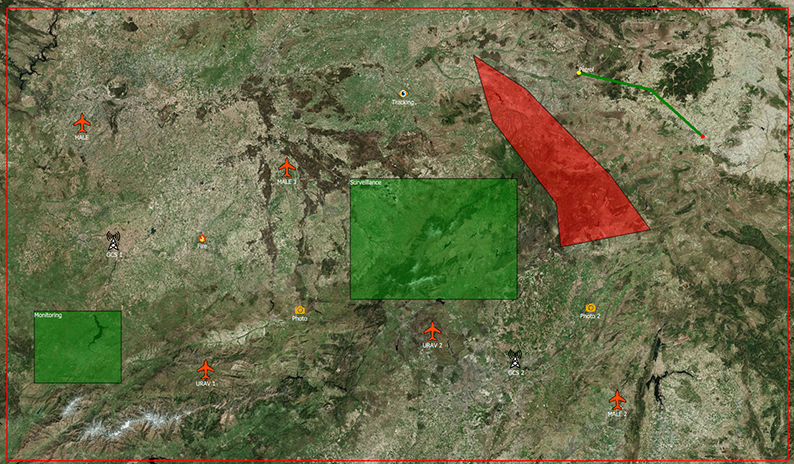}
        \caption{Mission 4.}
        \label{fig:d14}
    \end{subfigure}
    \quad
    \begin{subfigure}[b]{0.31\textwidth}
        \includegraphics[width=\textwidth]{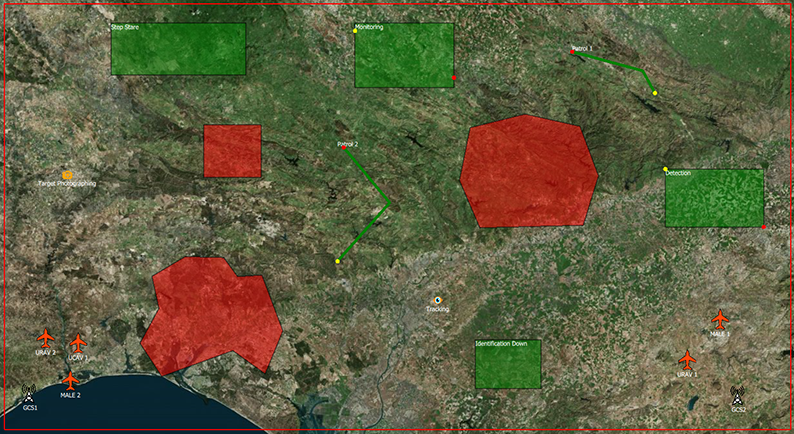}
        \caption{Mission 5.}
        \label{fig:d21}
    \end{subfigure}
    \quad
    \begin{subfigure}[b]{0.31\textwidth}
        \includegraphics[width=\textwidth]{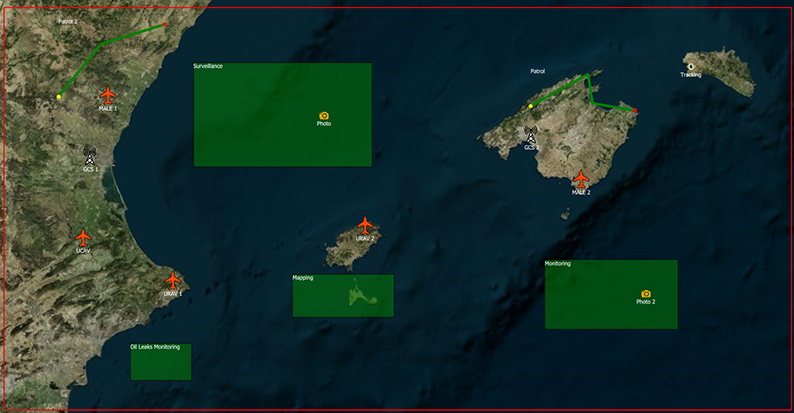}
        \caption{Mission 6.}
        \label{fig:d22}
    \end{subfigure}
    \caption{Mission Scenarios considered.}\label{fig:mission-scenario}
\end{figure*}

\subsection{Experimental setup}
In these experiments, 6 missions of increasing complexity have been designed to test the performance of the algorithm. These missions are described in Table \ref{tab:datasets}, showing the number of tasks, \glspl{uav}, \glspl{gcs}, \glspl{nfz} and task dependencies for each mission. Figure \ref{fig:mission-scenario} shows the scenario of each mission, where the green zones represent tasks and the red zones represent \glspl{nfz}. There are also some point tasks represented with an icon, such as photographing, tracking or fire extinguishing.

\begin{table}[!h]
\caption{Features of the different missions designed.}
\label{tab:datasets}
\begin{center}
\scalebox{0.72}{
\begin{tabular}{ccccccc }
  \toprule
  \begin{minipage}[t]{1cm}
  \centering
  Mission \\
  Id.
  \end{minipage} & Tasks & \begin{minipage}[t]{1.8cm}
  \centering
  Multi-UAV \\
  Tasks
  \end{minipage} & UAVs & GCSs & NFZs & \begin{minipage}[t]{2cm}
  \centering
  Time \\
  Dependencies
  \end{minipage}\\[3ex]
  \midrule
  1 & 5 & 0 & 3 & 1 & 0 & 0 \\
  2 & 6 & 1 & 3 & 1 & 1 & 0 \\
  3 & 6 & 1 & 4 & 2 & 2 & 1 \\
  4 & 7 & 1 & 5 & 2 & 1 & 2 \\
  5 & 8 & 2 & 5 & 2 & 3 & 1 \\
  6 & 9 & 2 & 5 & 2 & 0 & 2 \\
  \bottomrule
\end{tabular}
}
\end{center}
\end{table}

In the experimental phase, the number of solutions, the number of generations needed to converge, the runtime spent and the hypervolume of the solutions obtained are extracted. Every experiment is run 30 times, and the mean and standard deviation for every metric are computed.

Every execution of \gls{nsga2} use a selection criteria $\mu + \lambda = 10 + 100$, where $\lambda$ is the number of offspring (population size), and $\mu$ the elitism size (i.e. the number of the best parents that survive from current generation to the next). The mutation probability is $10\%$, the maximum number of generations is $300$ and the number of generations for the stopping criteria is $10$. The experiments have been run in a Intel Xeon CPU E5-2650 v3 2.30GHz with 10 cores and 250GB DDR4 RAM.

\subsection{Experiment with different variables}
In this first experiment, we consider five approaches with different variables of the \gls{csp}:

\begin{enumerate}
\item Only task assignments. In this approach, only task assignments and orders are considered in the encoding. In addition, order and temporal constraints are considered.
\item Using Flight Profiles. This approach extends the first one by adding flight profile variables and constraints.
\item Using Sensors. This approach extends the first one by adding sensor variables and constraints.
\item Using \glspl{gcs}. This approach extends the first one by adding \gls{gcs} variables and constraints.
\item Complete encoding. This approach comprises all the previous ones, including all the variables of \gls{csp} model for the \gls{mpp}.
\end{enumerate}

\begin{figure*}[hbt]
\centering
\includegraphics[width=\textwidth]{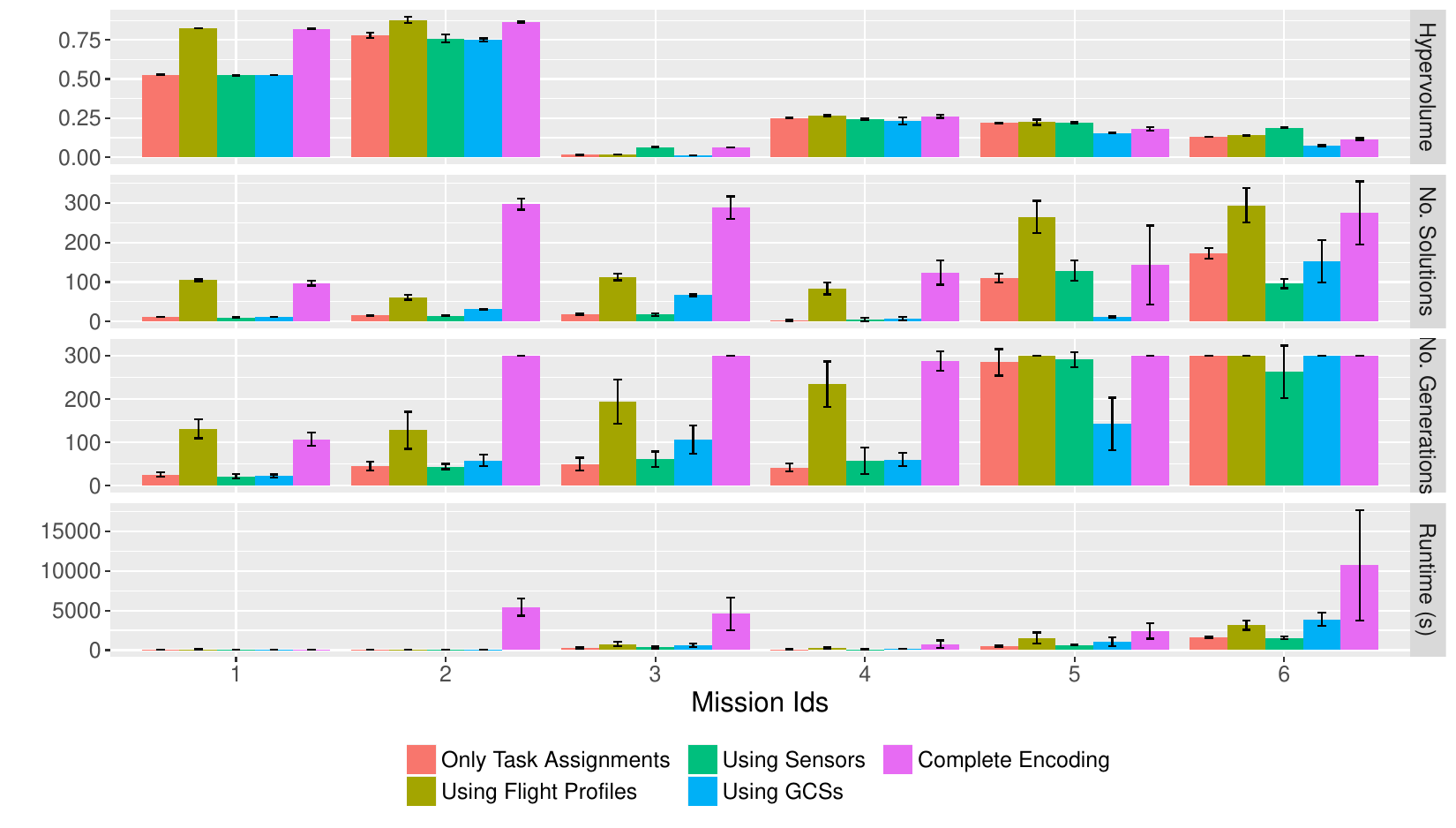}
\caption{Hypervolume, number of solutions, number of generations and runtime obtained with NSGA-II for MPPs considering different approaches using different variables of the \gls{csp}.}
\label{fig:variableResults}
\end{figure*}

Each approach is executed with \gls{nsga2}, and the results obtained are presented in Figure \ref{fig:variableResults}.

This figure presents bar graphs with the mean and error bars with the standard deviation for the every metric considered. It can be noticed that for the first missions, the flight profile is most significant than the sensors and the \glspl{gcs}, as more solutions are obtained and more runtime is spent. Obviously, the complete encoding presents the higher number of solutions, hypervolume and runtime. As the complete encoding approach considers much more variables and constraints than the rest, the runtime is always higher than any other approach.

For the most complex missions, it can be seen that the problems do not converge within the 300 generations considered. This can also be noticed as the standard deviation is quite large, specially for the number of solutions and the runtime.

The main conclusion that can be extracted here is that adding variables makes the complexity of the problem grow exponentially, and it becomes quite hard for the algorithm to find the \gls{pof}.

\subsection{Experiment with increasing number of constraints}
The second experiment consists of four approaches with increasing number of constraints of the \gls{csp}:

\begin{enumerate}
\item Basic constraints. This approach considers the complete encoding approach of the previous experiment, where temporal, order, \gls{gcs}, sensor and flight profile constraints are considered.
\item Adding checking constraints. This approach adds to the first one the checking constraints, i.e. flight time, fuel and distance constraints.
\item Adding path constraints. This approach adds to the previous one the path constraints, including the climb and descent consideration, the \gls{nfz} avoidance and the distance to ground computation.
\item Adding LOS. This approach adds the \gls{los} constraints to the previous one, resulting in the complete \gls{csp} model for \gls{mpp} defined in section \ref{uavmpp}.
\end{enumerate}

\begin{figure*}[hbt]
\centering
\includegraphics[width=\textwidth]{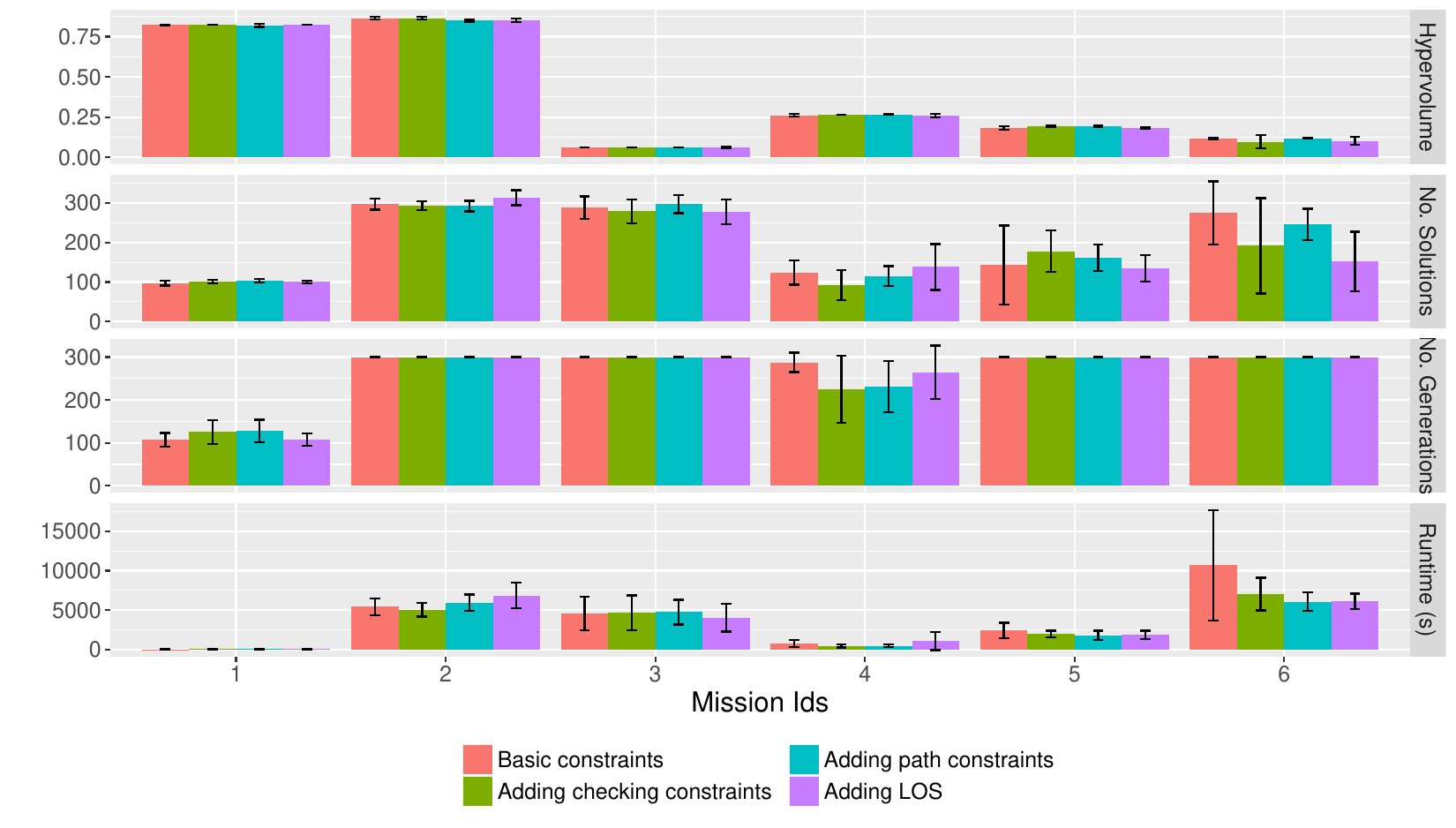}
\caption{Hypervolume, number of solutions, number of generations and runtime obtained with NSGA-II for MPPs considering different approaches with increasing number of constraints of the \gls{csp}.}
\label{fig:constraintResults}
\end{figure*}

Each approach is executed with \gls{nsga2}, and the results obtained are presented in Figure \ref{fig:constraintResults}.

These results show that adding more constraints do no affect the less complex missions, as the results obtained for all metrics is similar.

On the other hand, for the most complex mission, it is appreciable that the number of solutions and the runtime decrease as the number of constraints increase. This is due to many solutions becoming invalid due to the new constraints.

It is curious that the complexity of the path and \gls{los} constraints, which should highly increase the runtime of the algorithm, is not appreciated in the results (only Mission 2 shows a low increase in the runtime). This is because valid solutions take much more runtime to be checked, but as the number of solutions has been highly reduced, most solutions checked are invalid, which do not take so much time to be checked.

\subsection{Experiment with different constraints in the GA}
In this last experiment, we consider five approaches related with the consideration of constraints in the \gls{ga} process (see section \ref{gaconstraints}):

\begin{enumerate}
\item CSP Penalty Fitness. In this approach, no constraint is considered in the \gls{ga} setup nor the genetic operators, so the complete \gls{csp} model works as a penalty function.
\item Sensor constraints in GA. This approach considers the sensor constraints in the setup and mutation of \gls{nsga2}, constraining the generation of solutions to those with valid sensors.
\item Dependency constraints in GA. This approach considers dependency constraints in the setup and genetic operators of \gls{nsga2}, so a solution not fulfilling the task dependencies of the problem will not be generated.
\item GCS constraints in GA. This approach considers the \gls{gcs} constraints in the setup and genetic operators of the algorithm, avoiding invalid types of vehicles and surpassing of the maximum number of \glspl{uav} supported by the \gls{gcs}.
\item Sensor, GCS and Dependency constraints in GA. This approaches combines the three types of constraints considered in the previous approaches.
\end{enumerate}

All these approaches are run 30 times by the algorithm presented in section \ref{moeaapproach}, and the results obtained are presented in Figure \ref{fig:constraintResults}, representing the mean and standard deviation for each metric considered.

\begin{figure*}[hbt]
\centering
\includegraphics[width=\textwidth]{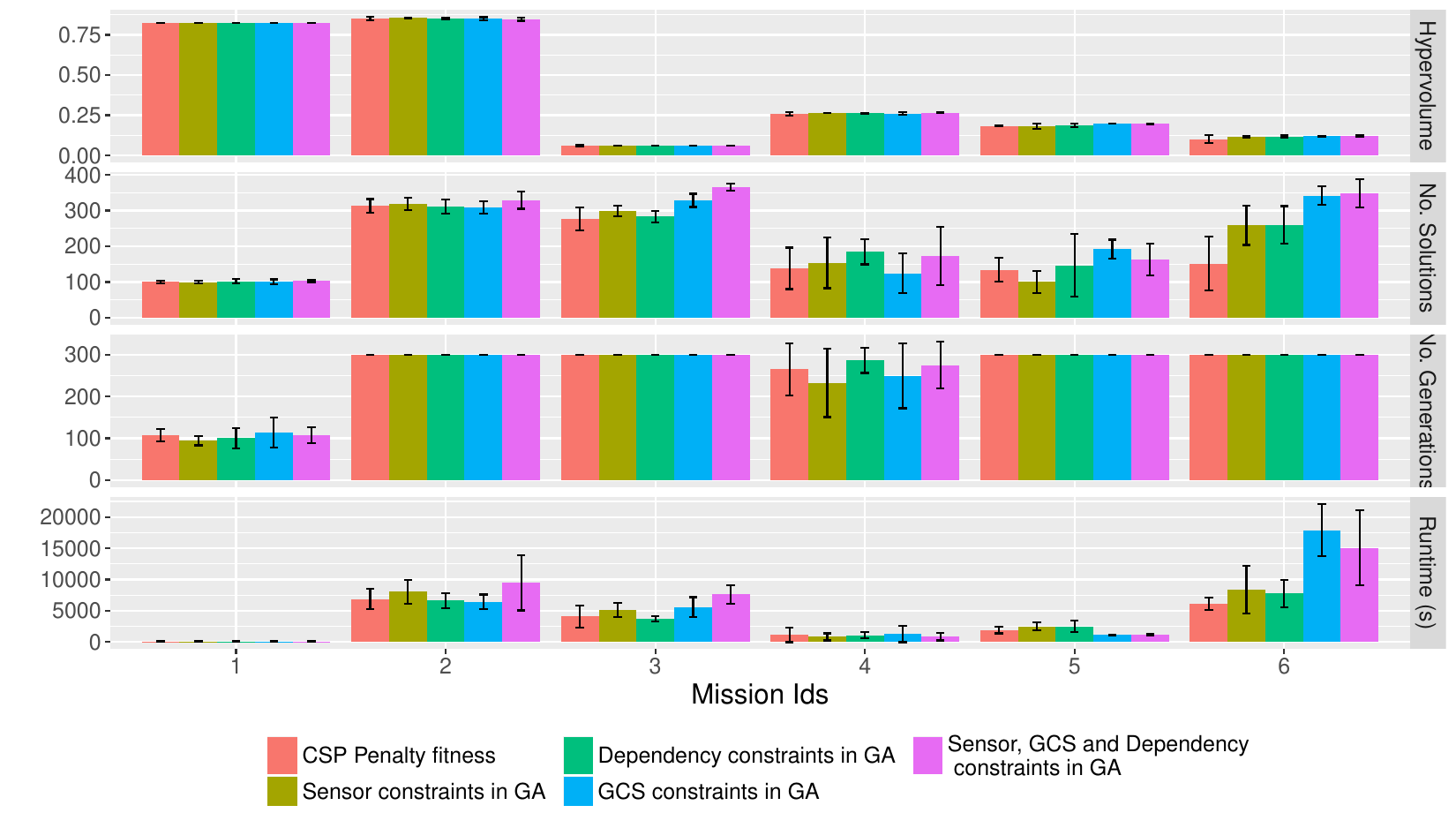}
\caption{Hypervolume, number of solutions, number of generations and runtime obtained with NSGA-II for MPPs considering different approaches using constraints in the \gls{ga} process.}
\label{fig:gaResults}
\end{figure*}

The results obtained show that the number of solutions obtained increase when adding constraints to the \gls{ga} process, specially for the most complex missions. As the algorithm reached its maximum number of generations, it is clear that using these constraints in the \gls{ga} setup accelerates the search of valid solutions, leading to a higher number of solutions in the end.

As can be seen, the runtime highly increases when using these constraints are used. This is because these approaches, specially the last one, find more valid solutions, which take much more time to be checked by the \gls{csp} model than the invalid ones.

Regarding at concrete constraints, it cannot be concluded if some constraint helps more than other in the search of valid solutions, as each mission got better results with different constraints. The less complex mission got better results just using sensor constraints, mission 4 got better results with dependency constraints, and the most complex obtained better results with \gls{gcs} constraints.

\begin{figure*}[!t]
     \centering
     \begin{subfigure}[b]{0.31\textwidth}
         \includegraphics[width=\textwidth]{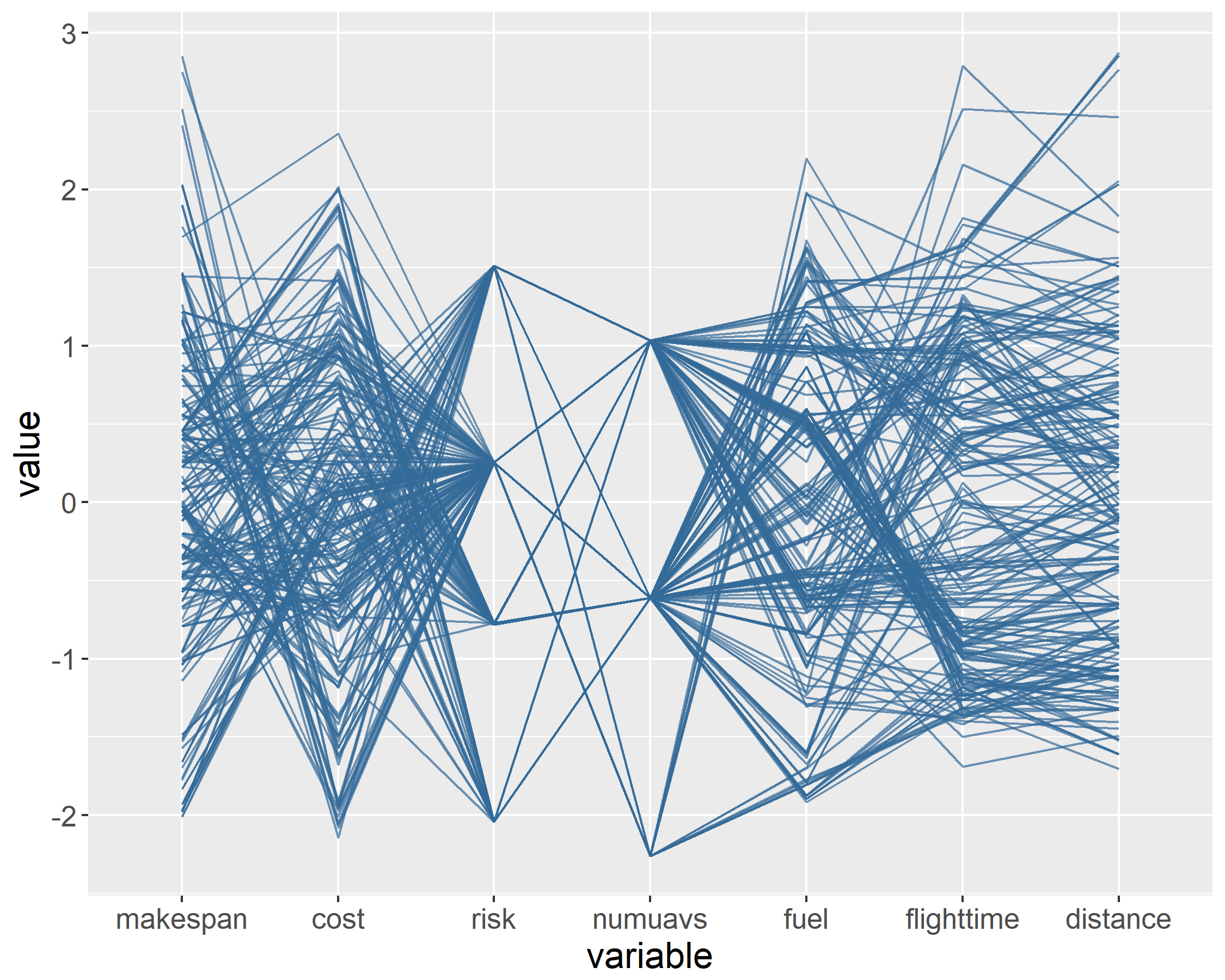}
         \caption{CSP Penalty Fitness.}
         \label{fig:parallel_csp_penalty}
     \end{subfigure}
     \quad
     \begin{subfigure}[b]{0.31\textwidth}
         \includegraphics[width=\textwidth]{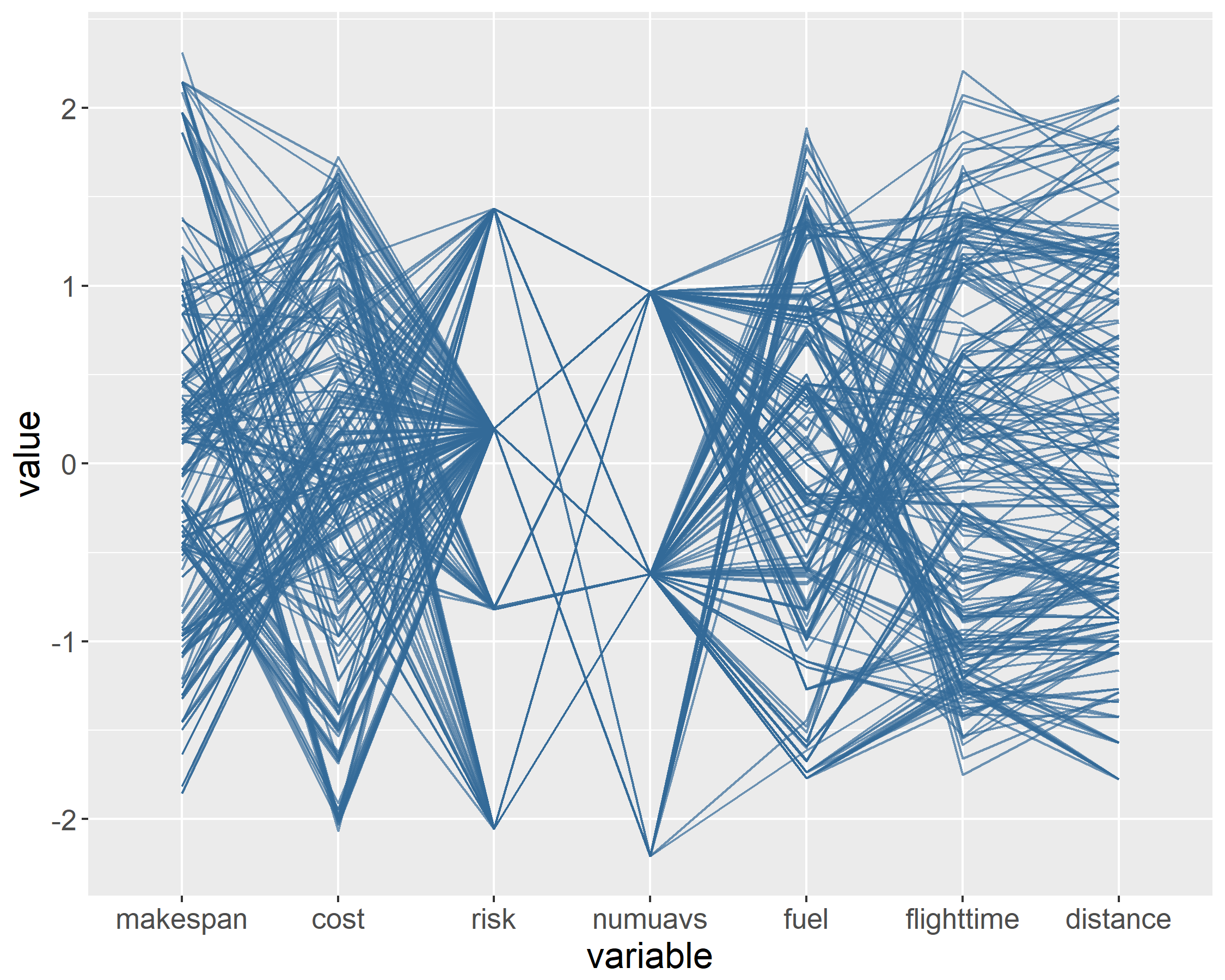}
         \caption{Sensor constraints in GA.}
         \label{fig:parallel_sensor_ga}
     \end{subfigure}
     \quad
     \begin{subfigure}[b]{0.31\textwidth}
         \includegraphics[width=\textwidth]{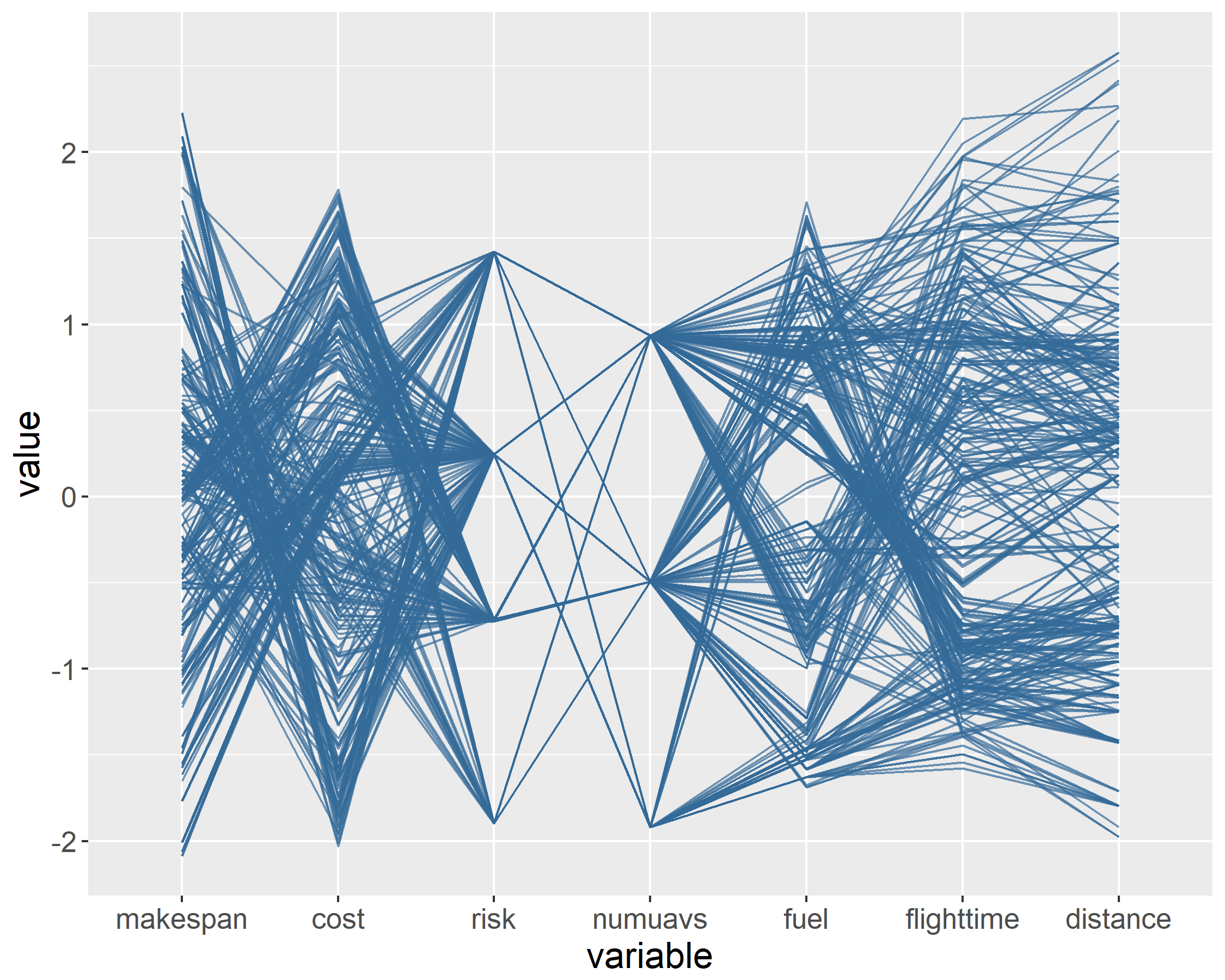}
         \caption{Dependency constraints in GA.}
         \label{fig:parallel_dependency_ga}
     \end{subfigure}
     \quad
     \begin{subfigure}[b]{0.31\textwidth}
         \includegraphics[width=\textwidth]{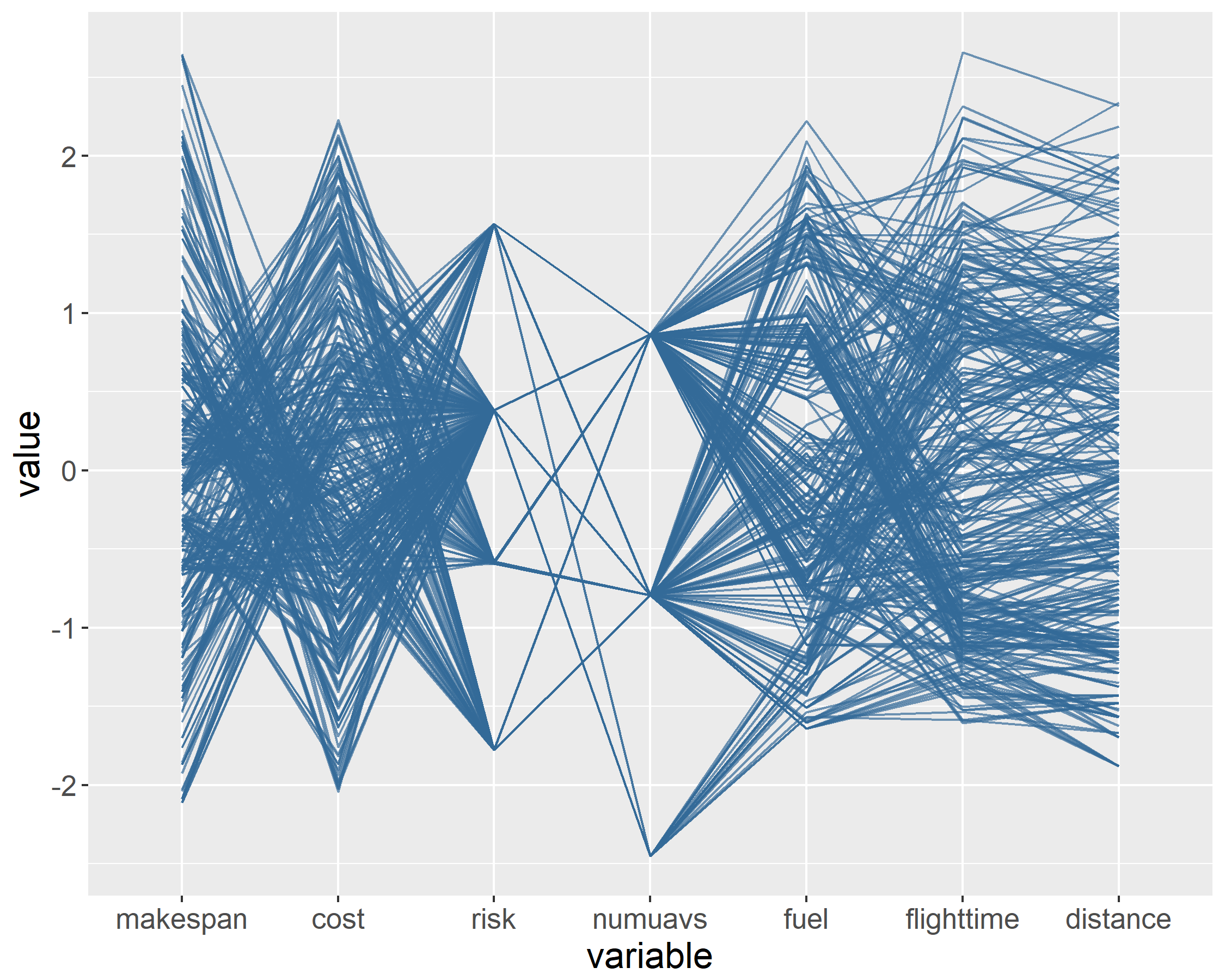}
         \caption{GCS constraints in GA.}
         \label{fig:parallel_gcs_ga}
     \end{subfigure}
     \quad
     \begin{subfigure}[b]{0.31\textwidth}
         \includegraphics[width=\textwidth]{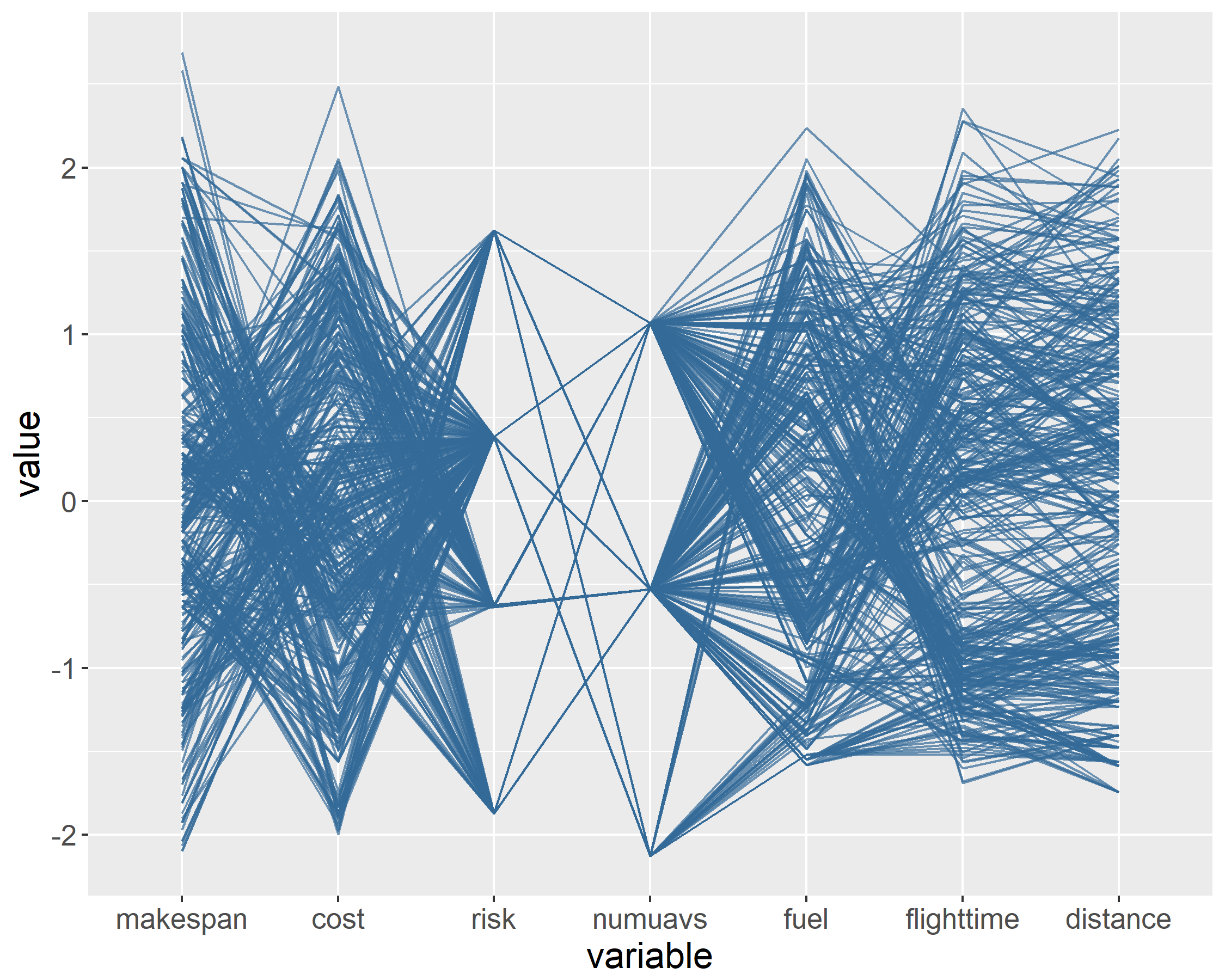}
         \caption{Sensor, dependency and GCS constraints in GA.}
         \label{fig:parallel_all_ga}
     \end{subfigure}
     \caption{Parallel Visualization Plots of solutions obtained in Mission 6 with NSGA-II-CSP approach considering all constraints as penalty in the fitness, or using sensor, dependency and GCS constraints in the generation of individuals in the GA.}\label{fig:parallel_ga}
 \end{figure*}

On the other hand, it is also appreciable that, although the number of solutions increase, the hypervolume obtained when using constraints in the \gls{ga} setup remains pretty similar to the one in the \gls{csp} penalty fitness approach. To study this, the solutions obtained for mission 6 have been plotted using a parallel visualization plot (see Figure \ref{fig:parallel_ga}). In order to ease the visualization and interpretation of the interplay between the different objectives, z-scores are used in these parallel plots, so changes are smoother depending on the standard deviation of the values for each objective.

In this representation it can be clearly seen, comparing Figure \ref{fig:parallel_csp_penalty} with the rest, that the new solutions appearing when using constraints in the \gls{ga} setup are close to the already obtained in the first approach. These solutions are similar to other solutions already computed in the first approach that just slightly increases one objective and slightly reduces another one. With this, the new solutions contribute very little to the hypervolume.

With this, it can be said that using constraints in the \gls{ga} setup and operators helps in the convergence of the \gls{pof}. But this \gls{pof} becomes pretty large with complex problems, so it is difficult to obtain it entirely with the current approach; and sometimes unnecessary as the new solutions obtained are not very relevant.

\section{Discussion}\label{conclusions}
In this work, the Multi-UAV \gls{mpp} has been presented and modelled as a \gls{csp}. Different variables and constraints of the \gls{csp} model defined the high complexity of this problem.

To solve this problem, a hybrid \gls{moea}-\gls{csp} algorithm was presented for the \gls{mpp}, where seven objectives of the problem must be minimized. \Gls{nsga2} was extended to deal with constraints both in the fitness function and some of them in the setup and genetic operators, aimed to reduce the search space of the problem. 

Three experiments were performed using varied datasets of different complexity. First, a comparative assessment of the variables is performed using different approaches of the problem using some specific variables. This experiment showed that the problem scales quite fast with the number of variables, becoming quite much difficult to optimize the problem with all the variables.

The second experiment used different approaches with an increasing number of constraints. Here, it was proved that the constraints reduce the space of valid solutions, making it faster.

Finally, the third experiment proved that adding constraints to the setup and genetic operators of the algorithm to reduce the search space of the algorithm, makes it converge better.

Nevertheless, it was concluded that as the complexity of the mission increases, the number of solutions in the \gls{pof} become huge. This implies that the time needed to obtain the complete\gls{pof} becomes intractable. On the other hand, in this problem the operator will in the end select one concrete solution to be executed, so getting so many solutions will highly increase the decision making process.

In future works, some new methods will be applied to outperform the convergence of the algorithm. In addition, the Multi-Objective approach will be changed to focus the search on the most significant solutions, avoiding unnecessary solutions that are not likely to be selected by the operator.

Finally, a \gls{dss} will be provided to rank and filter the solutions obtained in order to help the operator decide the final solution to be executed.

\section*{Compliance with Ethical Standards}
\textbf{Funding}: This study was funded by Spanish Ministry of Science and Education and Competitivity and European Regional Development Fund FEDER (TIN2014-56494-C4-4-P and TIN2017-85727-C4-3-P), Comunidad Aut\'onoma de Madrid (CIBERDINE S2013/ICE-3095) and Airbus Defence \& Space under Savier Project (FUAM-076915).

The Authors: Cristian Ramirez-Atencia and David Camacho declare that they have no conflict of interest.

\textbf{Ethical approval}: This article does not contain any studies with human participants or animals performed by any of the authors.

\begin{acknowledgements}
The authors would like to acknowledge the support obtained from Airbus Defence \& Space, specially from Savier Open Innovation project members: Jos\'e Insenser, C\'esar Castro, Gemma Blasco and In\'es Moreno.

\end{acknowledgements}

\bibliographystyle{spmpsci}      
\bibliography{biblio}   

\end{document}